\documentclass{article}

\usepackage{arxiv}

\usepackage[utf8]{inputenc} 
\usepackage[T1]{fontenc}    
\usepackage{hyperref}       
\usepackage{url}            
\usepackage{booktabs}       
\usepackage{amsfonts}       
\usepackage{nicefrac}       
\usepackage{microtype}      
\usepackage{lipsum}
\usepackage{graphicx}
\usepackage{xcolor}
\usepackage[percent]{overpic}
\usepackage{pict2e}

\title{Hierarchical multi-scale attention for semantic segmentation}

\author{
  Andrew Tao \\
  Nvidia\\
   \And
 Karan Sapra \\
 Nvidia\\
   \And
 Bryan Catanzaro \\
 Nvidia\\
}

\begin{document}
\maketitle

\begin{abstract}
Multi-scale inference is commonly used to improve the results of semantic segmentation. Multiple images scales are passed through a network and then the results are combined with averaging or max pooling. In this work, we present an attention-based approach to combining multi-scale predictions. We show that predictions at certain scales are better at resolving particular failures modes, and that the network learns to favor those scales for such cases in order to generate better predictions. Our attention mechanism is hierarchical, which enables it to be roughly $4$x more memory efficient to train than other recent approaches. In addition to enabling faster training, this allows us to train with larger crop sizes which leads to greater model accuracy. We demonstrate the result of our method on two datasets: Cityscapes and Mapillary Vistas. For Cityscapes, which has a large number of weakly labelled images, we also leverage auto-labelling to improve generalization. Using our approach we achieve a new state-of-the-art results in both Mapillary (61.1 IOU val) and Cityscapes (85.1 IOU test). 
\end{abstract}


\section{Introduction}
\begin{figure*}
\centering
\begin{tabular}{ccc}
Input images  & Prediction at 0.5x Scale &  Prediction at 2.0x Scale\\
\includegraphics[width=0.3\textwidth]{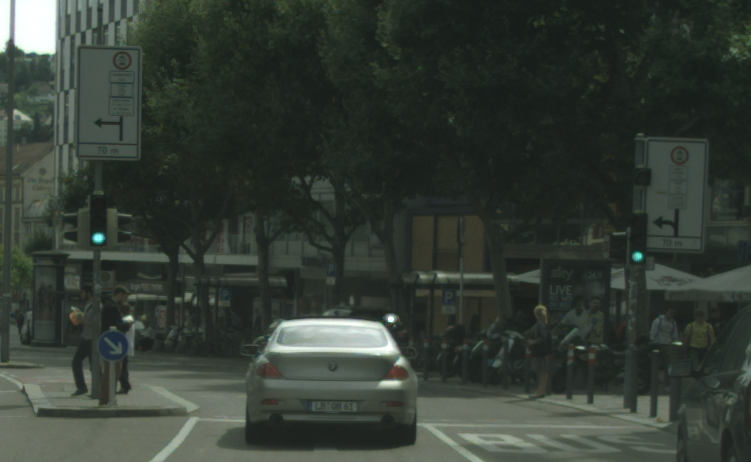} & 
\begin{overpic}[width=0.3\textwidth]{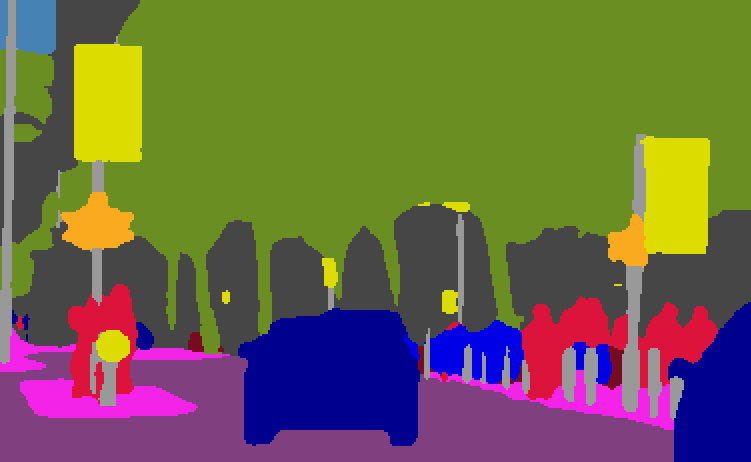}  
\put (67,15) {\color{red}\linethickness{0.5mm}\circle{25}}
\end{overpic}& 
\includegraphics[width=0.3\textwidth]{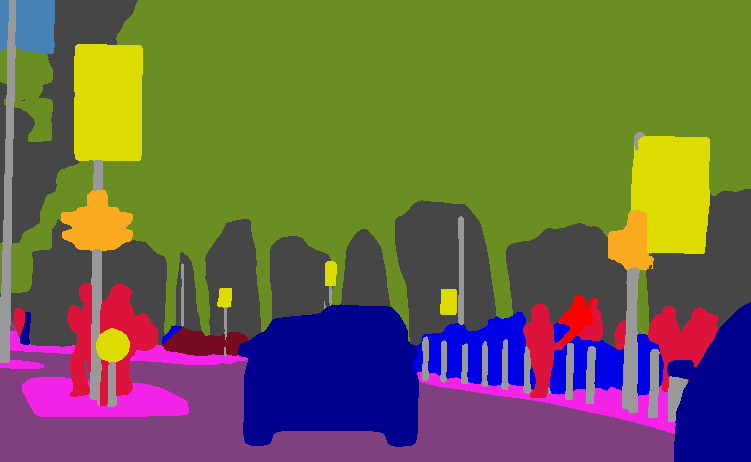} \\
\includegraphics[width=0.3\textwidth]{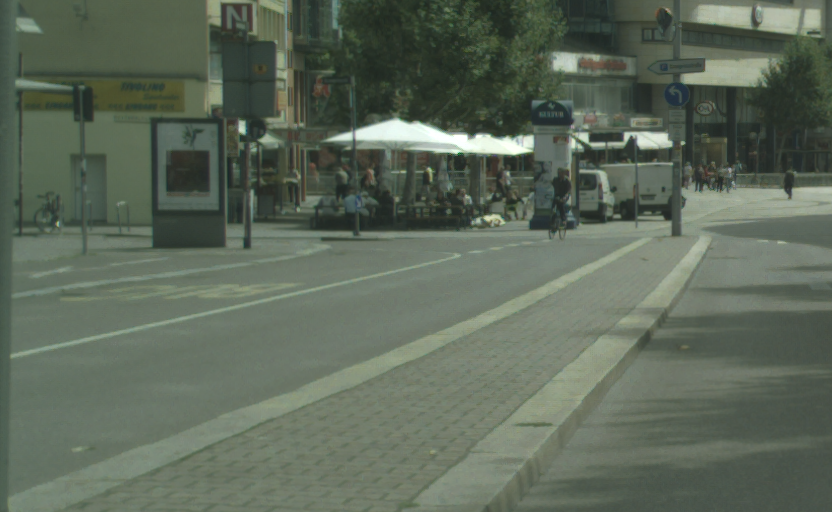} & 
\includegraphics[width=0.3\textwidth]{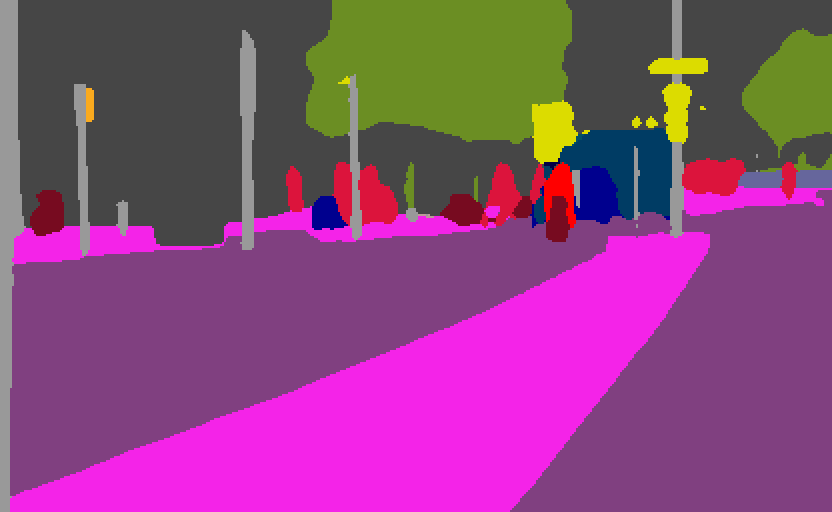} &
\begin{overpic}[width=0.3\textwidth]{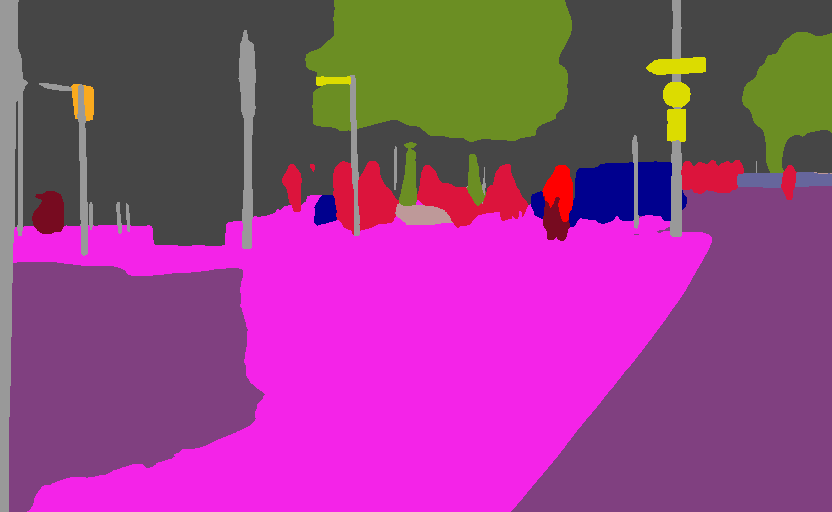}
\put (40,20) {\color{red}\linethickness{0.5mm}\circle{30}}
\end{overpic}
\end{tabular}

\caption{Illustration of common failures modes for semantic segmentation as they relate to inference scale. In the first row, the thin posts are inconsistently segmented in the scaled down (0.5x) image, but better predicted in the scaled-up (2.0x) image. In the second row, the large road / divider region is better segmented at lower resolution (0.5x).}
\label{fig:fig1}
\end{figure*}

The task of semantic segmentation is to label all pixels within an image as belonging to one of N classes. There is a trade off in this task in that certain types of predictions are best handled at lower inference resolution and other tasks better handled at higher inference resolution. Fine detail, such as the edges of objects or thin structures, is often better predicted with scaled up images sizes. And at the same time, predictions of large structures, which requires more global context, is often done better at scaled down image sizes, because the network's receptive field can observe more of the necessary context. We refer to this latter issue as class confusion. Examples of both of these cases are presented in Figure~\ref{fig:fig1}.

Using multi-scale inference is a common practice to address this trade off. Predictions are done at a range of scales, and the results are combined with averaging or max pooling. Using averaging to combine multiple scales generally improves results, but it suffers the problem of combining the best predictions with poorer ones. For example, if for a given pixel, the best prediction comes from the 2x scale, and a much worse prediction comes from the 0.5x scale, then averaging will combine these predictions, resulting in sub-par output. Max-pooling, on the other hand, selects only one of N scales to use for a given pixel, while the optimal answer may be a weighted combination across the different scales of predictions.

To address this problem, we adopt an attention mechanism to predict how to combine multi-scale predictions together at a pixel level, similar to the method proposed by Chen et. al.~\cite{chen2015attention}. We propose a hierarchical attention mechanism by which the network learns to predict a relative weighting between adjacent scales. In our method, because of it's hierarchical nature, we only require to augment the training pipeline with one extra scale whereas other methods such as \cite{chen2015attention} require each additional inference scale to be explicitly added during the training phase. For example, when the target inference scales for multi-scale evaluation are \{0.5, 1.0 and 2.0\}, other attention methods require the network to first be trained with all of those scales, resulting in $4.25$x ($0.5^2$ + $2.0^2$) extra training cost.  Our method only requires adding an extra 0.5x scale during training, which only adds $0.25$x ($0.5^2$) cost. Furthermore, our proposed hierarchical mechanism also provides the flexibility of choosing extra scales at inference time as compared to previous proposed methods that are limited to only use training scales during inference.

To achieve state-of-the-art results in Cityscapes, we also adopt an auto-labelling strategy of coarse images in order to increase the variance in the dataset, thereby improving generalization. Our strategy is motivated by multiple recent works, including ~\cite{xie2019selftraining,arazo2019pseudo,lee2013pseudo}. As opposed to the typical soft-labelling strategy, we adopt hard labelling in order to manage label storage size, which helps to improve training throughput by lowering the disk IO cost.

\subsection{Contributions}
\begin{itemize}
\item An efficient hierarchical multi-scale attention mechanism that helps with both class confusion and fine detail by allowing the network to learn how to best combine predictions from multiple inference scales
\item A hard-threshold based auto-labelling strategy which leverages unlabelled images and boosts IOU.
\item We achieve state-of-the-art results in Cityscapes (85.1 IOU) and Mapillary Vistas (61.1 IOU)
\end{itemize}

\section{Related Work}

\textbf{Multi-scale context methods.} State-of-the-art semantic segmentation networks use network trunks with low output stride. This allows the networks to be able to resolve fine detail better but it also has the effect of shrinking the receptive field. This reduction in the receptive field can cause networks to have difficulty with predicting large objects in a scene. Pyramid pooling can counteract the shrunken receptive field by assembling multi-scale context. PSPNet~\cite{zhao2017pyramid} use a spatial pyramid pooling module which assembles features at multiple scales using the features obtained from the final layer of network trunk using a sequence of pooling and convolution operation. DeepLab~\cite{chen2018encoder} uses Atrous Spatial Pyramid Pooling (ASPP) which employs atrous convolutions with different levels of dilation, thus creating, denser feature as compared to PSPNet. More recently, ZigZagNet~\cite{Lin_2019_CVPR} and ACNet ~\cite{fu2019adaptive} leverage intermediate features instead of just the features from the final layer of the network trunk to create the multi-scale context.

\textbf{Relational context methods.} In practice, pyramid pooling techniques attend to fixed, square context regions because pooling and dilation are typically employed in a symmetric fashion. Furthermore, such techniques tend to be static and not learned. However, relational context methods build context by attending to the relationship between pixels and are not bound to square regions. The learned nature of relational context methods allow context to be built based on image composition. Such techniques can build more appropriate context for non-square semantic regions, such as a long train or a tall thin lamp post.
OCRNet~\cite{yuan2019objectcontextual}, DANET~\cite{fu2018dual}, CFNet~\cite{zhang2019co}, OCNet~\cite{yuan2018ocnet} and other related work~\cite{A2Net,Zhang_2019_ICCV,chen2018graph,NIPS2018_7456,li2018beyond,NIPS2018_7886,li19,huang2018ccnet} use such relationships to build better context. 

\textbf{Multi-scale inference.} Both relation and multi-scale context methods ~\cite{chen2017rethinking,chen2018encoderdecoder,cheng2019panopticdeeplab,yuan2019objectcontextual} use multi-scale evaluation to achieve the best results. There are two common approaches to combining network predictions at multiple scales: average and max pooling, with average pooling being more common. However, average pooling involves equally weighting output from different scales, which may be sub-optimal. To address this issue ~\cite{chen2015attention, yang2018attention} use using attention to combination multiple scales. Chen et. al. \cite{chen2015attention} train an attention head across all scales simultaneously using final features from a neural network. While Chen  et. al. use attention from a specific layer, Yang et. al. \cite{yang2018attention} use a combination of features from different network layers to build better contextual information. However, both of the aforementioned methods share the trait that the network and attention heads are trained with a fixed set of scales. Only those scales may be used at run-time, else the network must be re-trained. We propose a hierarchical based attention mechanism that is agnostic to number of scales during inference time. Furthermore, we show that our proposed hierarchical attention mechanism not only improves performance over average-pooling, but also allows us to diagnostically visualize the importance of different scales for classes and scenes. Furthermore, our method is orthogonal to other attention or pyramid pooling methods such as ~\cite{chen2018encoderdecoder,sinha2019multiscale,lin2016refinenet,yuan2019objectcontextual,Huang_2019_ICCV,fu2018dual,li2018pyramid} as these methods use single scale image and perform attention to better combine multi-level features for generating high-resolution predictions.

\textbf{Auto-labelling.} Most recent semantic segmentation work for Cityscapes in particular has utilized the \char`\~$20,000$ coarsely labelled images as-is for training state-of-the-art models ~\cite{yuan2018ocnet, semantic_cvpr19}. However, a significant amount of each coarse image is unlabelled due to the coarseness of the labels. To achieve state-of-the-art results on Cityscapes, we adopt an auto-labelling strategy, motivated by Xie et. al.~\cite{xie2019selftraining}, other semi-supervised self-training in semantic segmentation~\cite{Lian_2019_Pyramid,Li_2019_bidirection,Luc2017futureSeg,Zou2018DAClassBalance,Zou_2019_CRST}, and other approaches based on pseudo label such as ~\cite{lee2013pseudo,iscen2019label,shi2018transductive,arazo2019pseudo}. We generate dense labels for the coarse images in Cityscapes. Our generated labels have very few unlabelled regions, and thus we are able to take advantage of the full content of the coarse images.

While most image classification auto-labelling work use continuous or \textit{soft} labels, we generate \textit{hard} thresholded labels, for storage efficiency and training speed. With soft labels, a teacher network provides a continuous probability for each of N classes for each pixel of an image, whereas for hard labels a threshold is used to pick a single top class per pixel. Similar to~\cite{li2019decoupled,lee2013pseudo} we generate hard dense labels for the coarse Cityscapes images. Examples are shown in Figure ~\ref{fig:fig4}. Unlike Xie. et. al.~\cite{xie2019selftraining}, we do not perform iterative refinement of our labels. Rather, we perform a single iteration of full training of our teacher model with the default coarse and fine labelled provided images. After this joint training, we perform \emph{auto-labelling} of the coarse images, which are then substituted in our teacher training recipe to obtain state-of-the-art test results. Using our pseudo generated hard labels in combination with our proposed hierarchical attention, we are able to obtain state-of-the-art results on Cityscapes.

\section{Hierarchical multi-scale attention}
\label{sec:headings}

\begin{figure*}
  \centering
    \begin{overpic}[width=1.0\textwidth,tics=2,trim={1cm 1cm 1cm 1cm},clip]{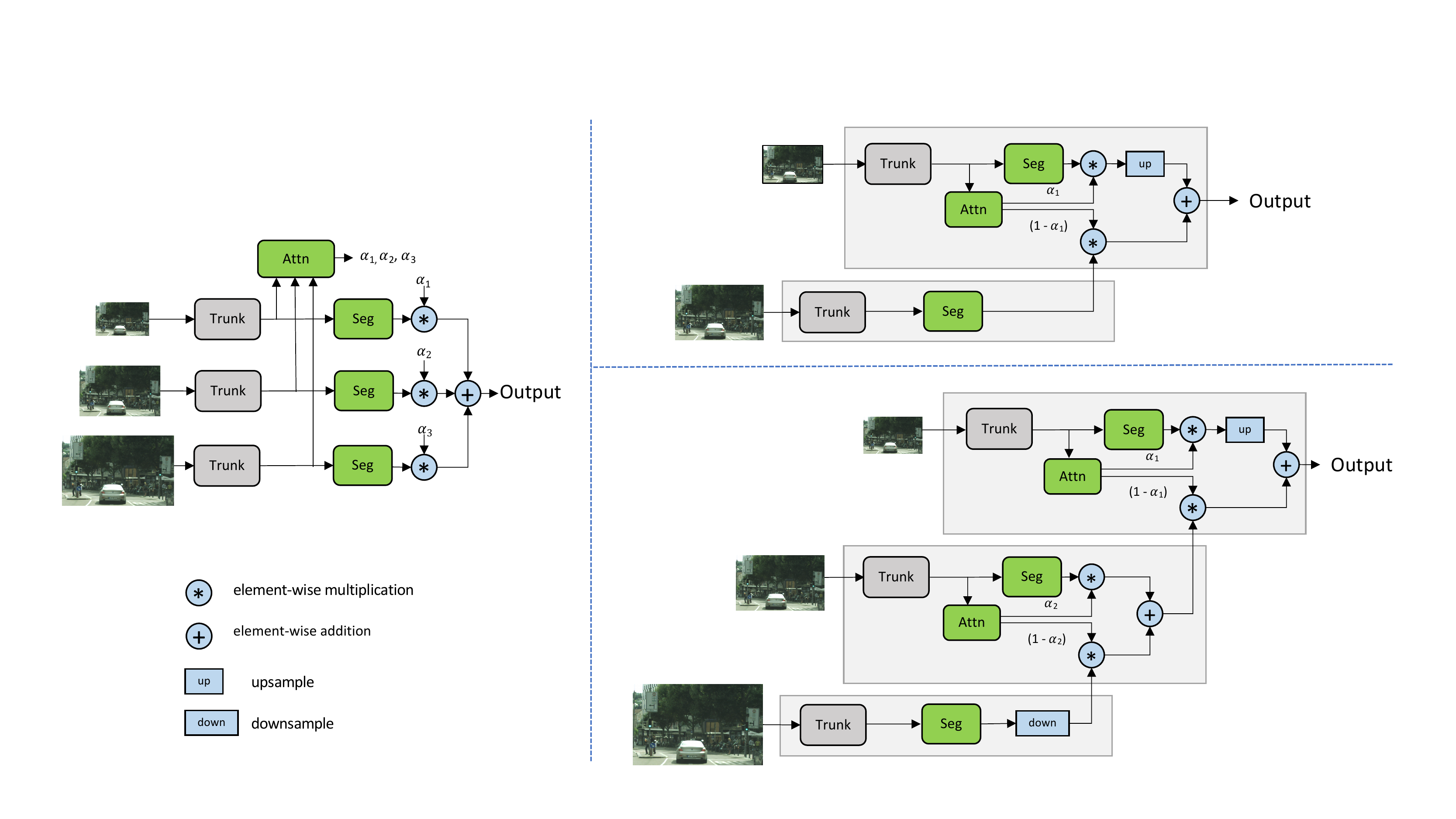}\qquad
    \put(12, 50){\small \textbf{Explicit Method}}
    \put(12, 18){\scriptsize \textbf{Training and Inference}}
    \put(60, 50){\small \textbf{Our Hierarchical Method}}
    \put(82, 32){\scriptsize \textbf{Training}}
    \put(82, 2){\scriptsize \textbf{Inference}}
    
    \put(46, 7){\tiny Scale 3}
    \put(52, 16.4){\tiny Scale 2}
    \put(60.5, 26.5){\tiny Scale 1}
    
    \put(47.6, 36.2){\tiny Scale 2}
    \put(52.9, 46.4){\tiny Scale 1}
    
    \put(4, 25){\tiny Scale 3}
    \put(4, 30.3){\tiny Scale 2}
    \put(4, 34.85){\tiny Scale 1}

    \end{overpic}
    \caption{\textbf{Network Architecture} Left and right panels show \textbf{explicit} vs. \textbf{hierarchical} (Ours) architectures, respectively. \textbf{Left} shows the architecture from ~\cite{chen2015attention}, where the attention for each scale is learned explicitly. \textbf{Right} shows our hierarchical attention architecture. \textbf{Right top} An illustration of our training pipeline, whereby the network learns to predict attention between adjacent scale pairs. \textbf{Right bottom} Inference is performed in a chained/hierarchical manner in order to combine multiple scales of predictions. Lower scale attention determines the contribution of the next higher scale.}
    
  \label{fig:fig2}
\end{figure*}
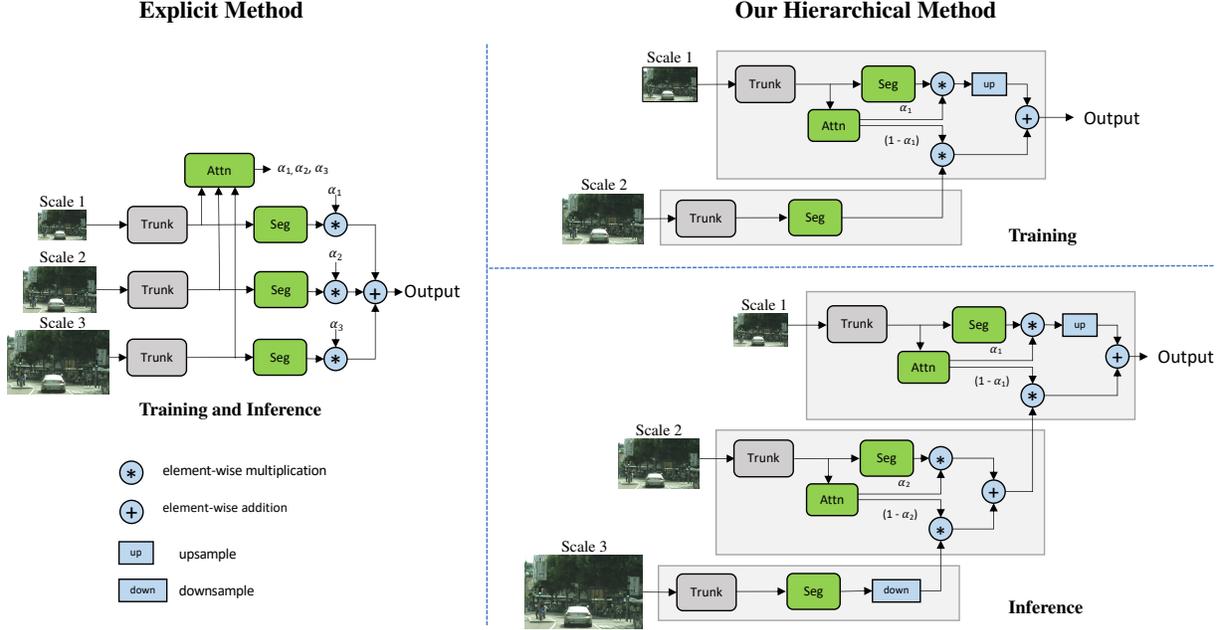

Our attention mechanism is conceptually very similar to that of~\cite{chen2015attention}, where a dense mask is learned for each scale, and these multi-scale predictions are combined by performing pixel-wise multiplication between masks with the predictions followed by pixel-wise summation among the different scales to obtain the final results, see Figure~\ref{fig:fig2}. We refer to Chen's method as \textbf{explicit}. With our \textbf{hierarchical} method, instead of learning all attention masks for each of a fixed set of scales, we learn a relative attention mask between adjacent scales. When training the network, we only train with adjacent scale pairs. As shown in Figure~\ref{fig:fig2}, given a set of image features from a single (lower) scale, we predict a dense pixel-wise the relative attention between the two image scales. In practice, to obtain the pair of scaled images, we take a single input image and scale it down by a factor of 2, such that we are left with a 1x scale input and an 0.5x scaled input, although any scale-down ratio could be selected. It is important to note that the network input itself is a re-scaled version of the original training images because we use image scale augmentation when we train. This allows the network network learns to predict relative attention for a range of image scales. When running inference, we can hierarchically apply the learned attention to combine N scales of predictions together, in a chain of computations as shown in Figure and described by equation below. We give precedence to Lower scales and work our way up to higher scales, with the idea that they have more global context and can choose where predictions need to be refined by higher scale predictions.

More formally, during training a given input image is scaled by factor $r$ where $r=0.5$ denotes a down-sampling by factor of 2, $r=2.0$ denotes upsampling by factor of 2, $r=1$ denotes no operation. For our training, we choose $r=0.5$ and $r=1.0$. The two images with $r=1$ and $r=0.5$ are then sent through the shared network trunk, which produces semantic logits $\mathcal{L}$ and also an attention mask$(\alpha)$ for each scale, which are used to combine the logits $\mathcal{L}$ between scales. Thus for two scale training and inference, with $\mathcal{U}$ being the bilinear upsampling operation, $*$  and $+$ are pixel-wise multiplication and addition respectively, the equation can be formalized as:

\begin{equation}
    \centering
   \mathcal{L}_{\boldmath{{(r=1)}}} = \mathcal{U}(\mathcal{L}_{(r=0.5)} * \alpha_{(r=0.5)}) + ((1-\mathcal{U}(\alpha_{(r=0.5)})) * \mathcal{L}_{(r=1)} )
\end{equation}
There are two advantages using our proposed strategy:
\begin{itemize}
 \item At inference time, we can now flexibly select scales, thus adding new scales such $0.25$x or $2.0$x to a model trained with $0.5$x and $1.0$x  is possible with our proposed attention mechanism chains together in a hierarchical way. This differs from previously proposed methods that limited to using the same scaled that were used during model training.
 
 \item This hierarchical structure allows us to improve on the training efficiency as compared to the explicit method. With the explicit method, if using scales $0.5$, $1.0$, $2.0$, the training cost is $0.5^2 + 1.0^2 + 2.0^2 = 5.25$, relative to single-scale training. With our hierarchical method the training cost is only $0.5^2 + 1.0^2 = 1.25$.
\end{itemize}
 
 \subsection{Architecture}
 \textbf{Backbone} For the ablation studies in this section, we use ResNet-50 \cite{he2016deep} (configured with output stride of 8) as the trunk for our network. For state-of-the-art results, we use a larger, more powerful trunk, HRNet-OCR \cite{yuan2019objectcontextual}. \textbf{Semantic Head}: Semantic predictions are performed by a dedicated fully convolutional head consisting of $($3x3 conv$)$ $\rightarrow$ $($BN$)$ $\rightarrow$ $($ReLU$)$  $\rightarrow$ $($3x3 conv$)$ $\rightarrow$ $($BN$)$ $\rightarrow$ $($ReLU$)$ $\rightarrow$ $($1x1 conv$)$. The final convolution outputs \textit{num\_classes} channels. \textbf{Attention Head}:Attention predictions are done using a separate head that is structurally identical to the semantic head, except for the final convolutional output, which outputs a single channel. When using ResNet-50 as the trunk, the semantic and attention heads are fed with features from the final stage of ResNet-50. When using HRNet-OCR, the semantic and attention heads are fed with features out of the OCR block. With HRNet-OCR, there also exists an \textbf{auxiliary semantic head}, which takes its features directly from the HRNet trunk, before OCR. This head consists of $($1x1 conv$)$ $\rightarrow$ $($BN$)$ $\rightarrow$ $($ReLU$)$ $\rightarrow$ $($1x1 conv$)$. After attention is applied to the semantic logits, the predictions are upsampled to the target image size with bilinear upsampling.

\begin{table}[htb!]
\centering
\begin{tabular}{|c|c|c|c|c|}
\hline
\textbf{Method}     & \textbf{Eval scales ($r$)} & \textbf{IOU} & \textbf{FLOPS (relative)} & \textbf{Minibatch training time (sec)}\\ \hline
Single Scale      & $1.0$         & $47.7$           & $1.00$x  & $0.80$ \\ \hline
AvgPool      & $0.5,1.0,2.0$         & $49.4$           & $1.00$x  & $0.80$ \\ \hline
AvgPool      & $0.25,0.5,1.0,2.0$     & $48.7$           & $1.00$x  & $0.80$ \\ \hline
Explicit   & $0.5,1.0,2.0$          & $51.4$           & $5.25$x  & $3.08$ \\ \hline
Hierarchical (Ours)         & $0.5,1.0,2.0$          & \textbf{51.6}  & $1.25$x & $1.17$ \\ \hline
Hierarchical (Ours)         & $0.25,0.5,1.0,2.0$     & \textbf{52.2}  & $1.25$x  & $1.17$\\ \hline
\end{tabular}

\caption{Comparison of our hierarchical multi-scale attention method vs. other approaches on Mapillary \textit{validation} set. The network architecture is DeepLab V3+ with a \textit{ResNet-50} trunk. \textbf{Eval scales}: scales used for multi-scale evaluation. \textbf{FLOPS}: the relative amount of flops consumed by the network for training. \textbf{Minibatch time}: measured training minibatch time on an Nvidia Tesla V100 GPU.}
\end{table}
\subsection{Analysis}

In order to evaluate the effectiveness of our multi-scale attention approach, we train networks with a DeepLab V3+ architecture and ResNet50 trunk. In Table 1, we show that our hierarchical attention approach results in better accuracy (51.6) as compared to the baseline averaging approach (49.4) or the explicit approach (51.4). We also observe significantly better results with our approach when adding the 0.25x scale Unlike the \textbf{explicit} method, our method does not require re-training the network when using the additional 0.25x scale. This \textbf{flexibility at inference time} is a key benefit of our method. We can train once but evaluate flexibly with a range of different scales.

\begin{figure}
\centering
\begin{tabular}{c}
Input images \\
\includegraphics[width=0.48\textwidth]{demo_crops/stuttgart_00_000000_000090_leftImg8bit.png} 
\includegraphics[width=0.48\textwidth]{demo_crops/stuttgart_00_000000_000272_leftImg8bit.png} \\
Semantic and Attention prediction at scale $0.5$x \\
\includegraphics[width=0.24\textwidth]{demo_crops/stuttgart_00_000000_000090_leftImg8bit_pred_05x.png}  \includegraphics[width=0.24\textwidth]{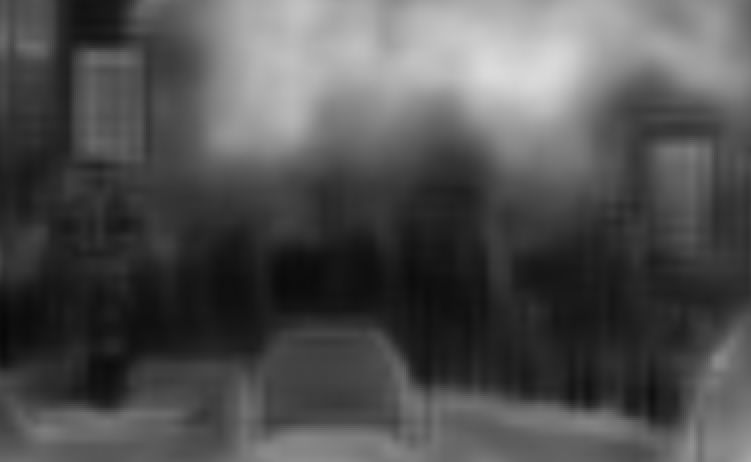} \includegraphics[width=0.24\textwidth]{demo_crops/stuttgart_00_000000_000272_leftImg8bit_pred_05x.png} \includegraphics[width=0.24\textwidth]{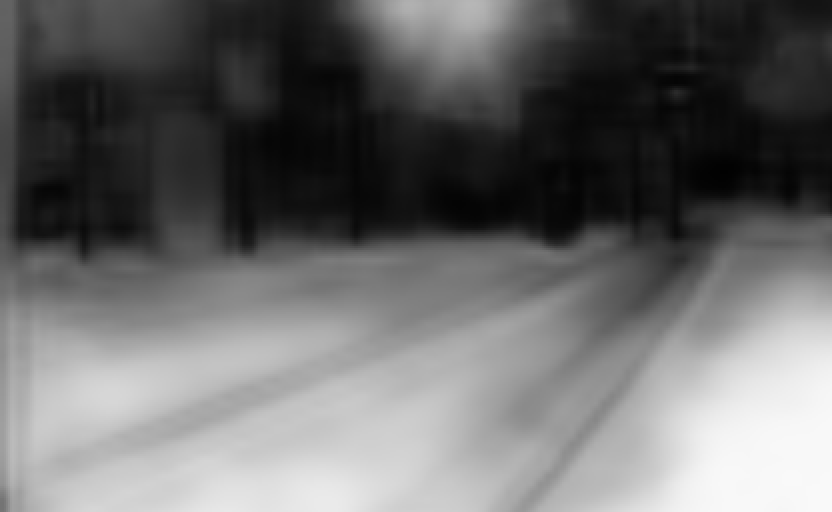}            \\
Semantic and Attention prediction at scale $1.0$x \\
\includegraphics[width=0.24\textwidth]{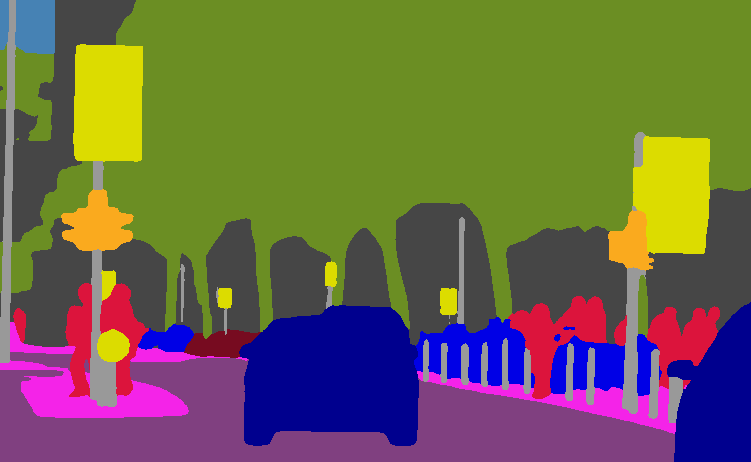}  \includegraphics[width=0.24\textwidth]{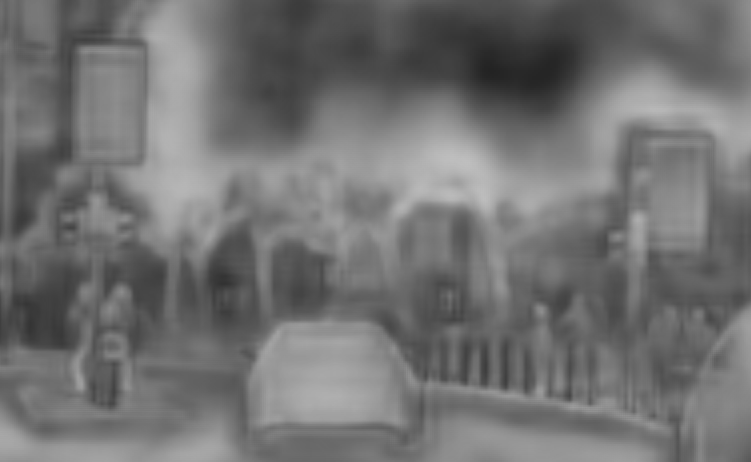}  \includegraphics[width=0.24\textwidth]{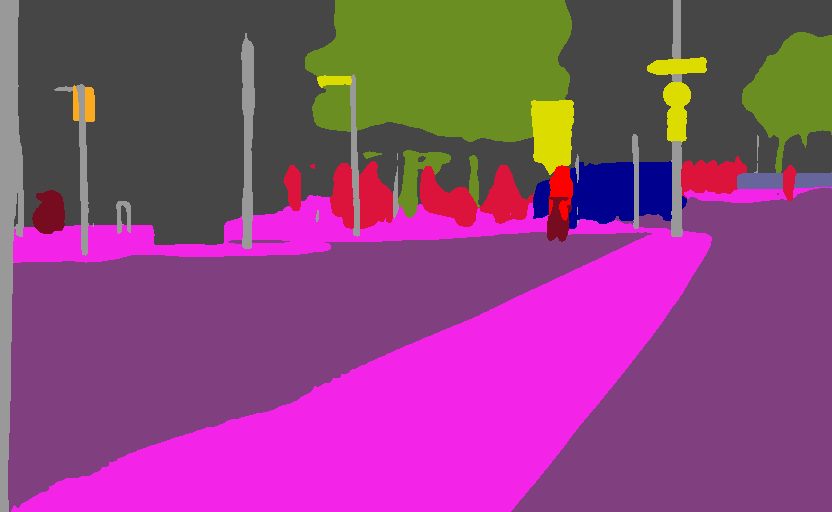} \includegraphics[width=0.24\textwidth]{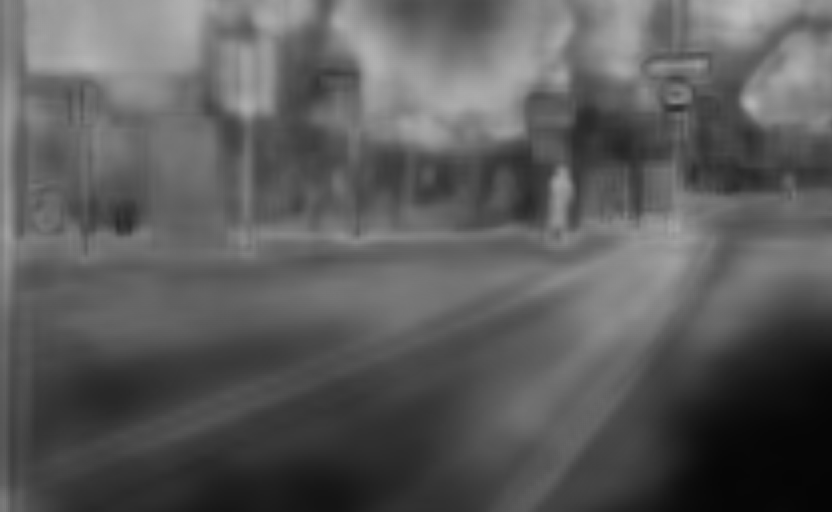}            \\
Semantic and Attention prediction at scale $2.0$x\\    
\includegraphics[width=0.24\textwidth]{demo_crops/stuttgart_00_000000_000090_leftImg8bit_pred_2x.png}  \includegraphics[width=0.24\textwidth]{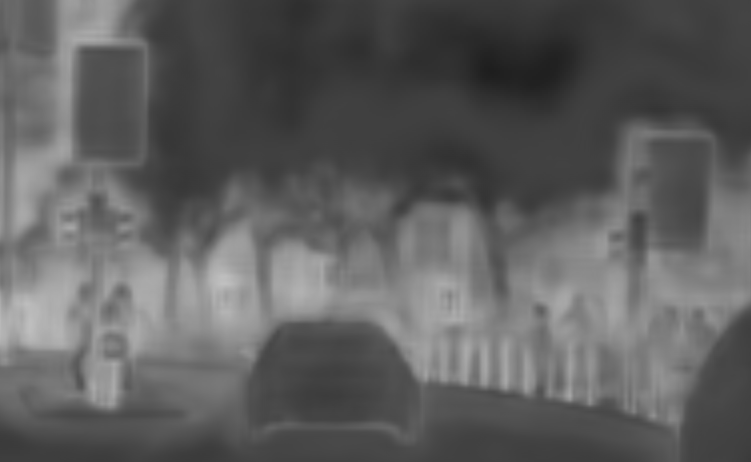}  \includegraphics[width=0.24\textwidth]{demo_crops/stuttgart_00_000000_000272_leftImg8bit_pred_2x.png} \includegraphics[width=0.24\textwidth]{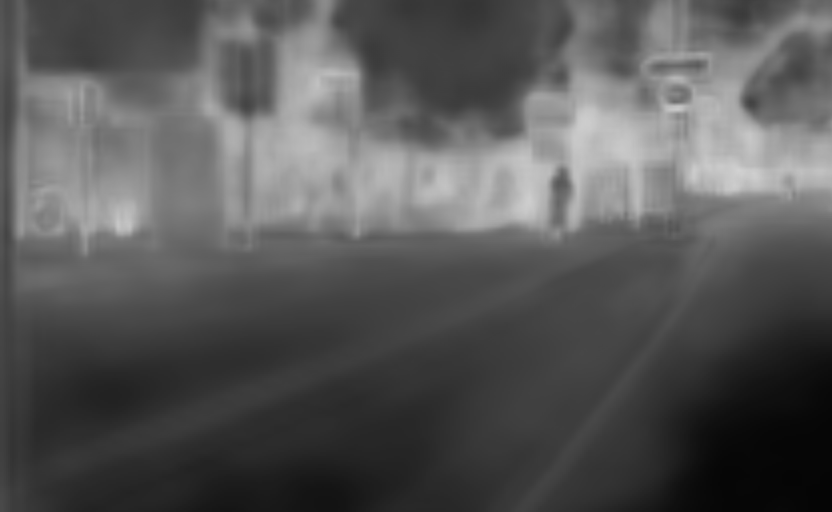}            \\
\end{tabular}

\caption{Semantic and attention predictions at every scale level for two different scenes. The scene on the \textbf{left} illustrates a fine detail problem while the scene on the \textbf{right} illustrates a large region segmentation problem. A white color for attention indicates a high value (close to 1.0). The attention values for a given pixel across all scales sums to 1.0. \textbf{Left}: The thin road-side posts are best resolved at 2x scale, and the attention successfully attends more to that scale than other scales, as evidenced by the white color for the posts in the 2x attention image. \textbf{Right}: The large road/divider region is best predicted at 0.5x scale, and the attention does successfully focus most heavily on the 0.5x scale for that region.}
    \label{fig:fig3}
\end{figure}

Furthermore, we also observe that with the baseline averaging multi-scale method, simply adding $0.25$x scale is detrimental to accuracy as it causes a $0.7$ reduction in IOU, whereas for our method, adding the extra 0.25x scale boosts accuracy by another $0.6$ IOU. With the baseline averaging method, the 0.25x prediction is so coarse that when averaged into the other scale, we observe classes such as lane marking, man-hole, phone-booth, street-light, traffic light and traffic sign (back and front ), bike racks, among others drop by $1.5$ IOU. The coarseness of the prediction hurts the edges and fine detail. However, with our proposed attention method, adding $0.25$x scale improves our result by $0.6$ since our network is able to apply the 0.25x prediction in the most appropriate way, staying away from using it around edges. Examples of this can be observed in Figure~\ref{fig:fig3}, where for the fine posts in the image on the left, very little of the posts are attended to by the 0.5x prediction, but a very strong attention signal is present in the 2.0x scale. Conversely, for the very large region on the right, the attention mechanism learns to most leverage the lower scale (0.5x) and very little of the erroneous 2.0x prediction.

\subsubsection{Single vs. dual-scale features}
While the architecture we settled upon feeds the attention head from features coming out of only the lower of two adjacent image scales (see Figure~\ref{fig:fig2}), we experimented with training the attention head with features from both adjacent scales. We did not observe significant difference in accuracy, so we settled on a single set of features.

\section{Auto Labelling on Cityscapes}

\begin{figure}
\large
\centering
\scalebox{0.75}{
\begin{tabular}{ccc}
\\
Original image & Original coarse label & Auto-generated coarse label\\
\includegraphics[width=0.4\textwidth]{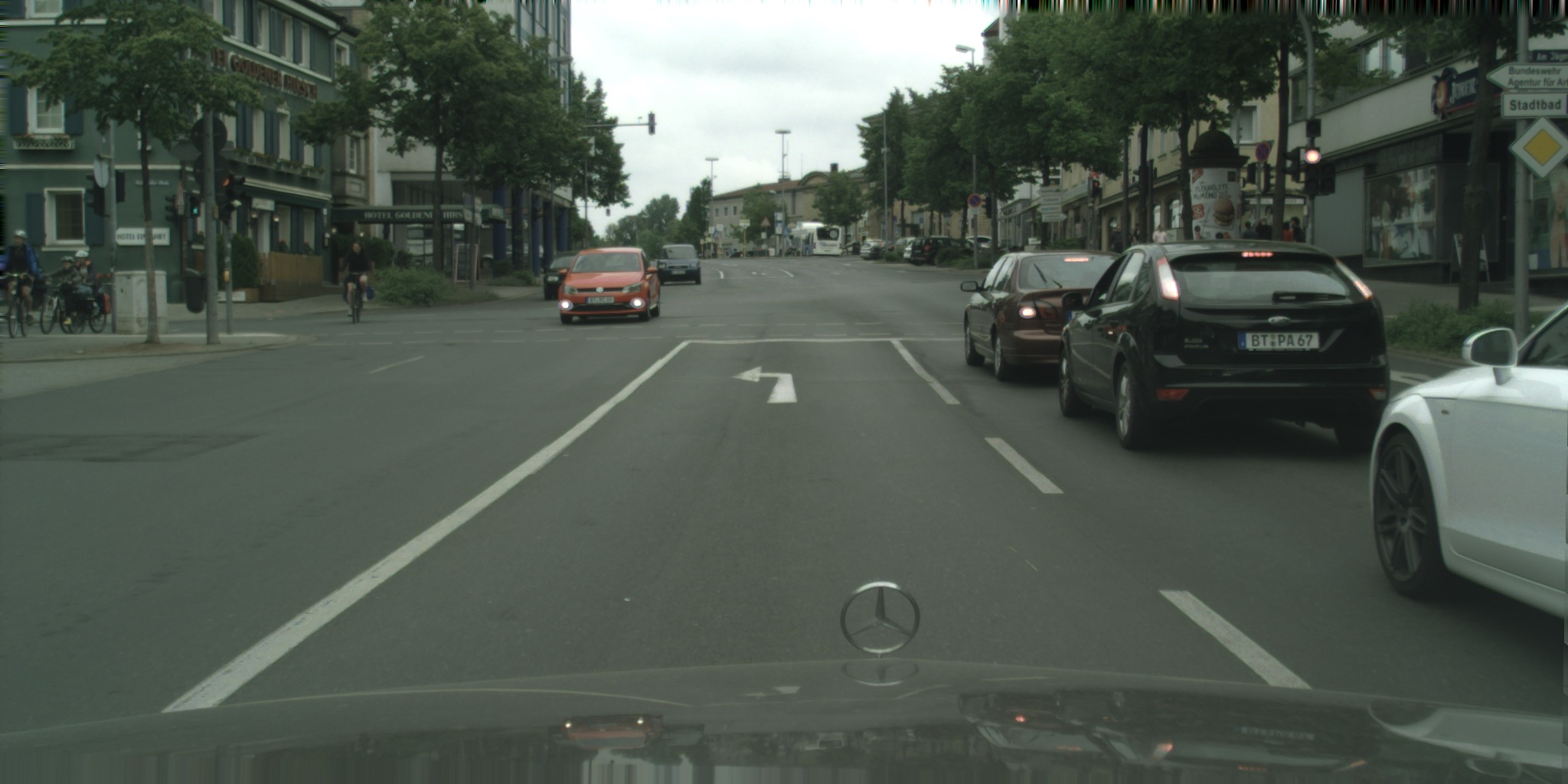} & \includegraphics[width=0.4\textwidth]{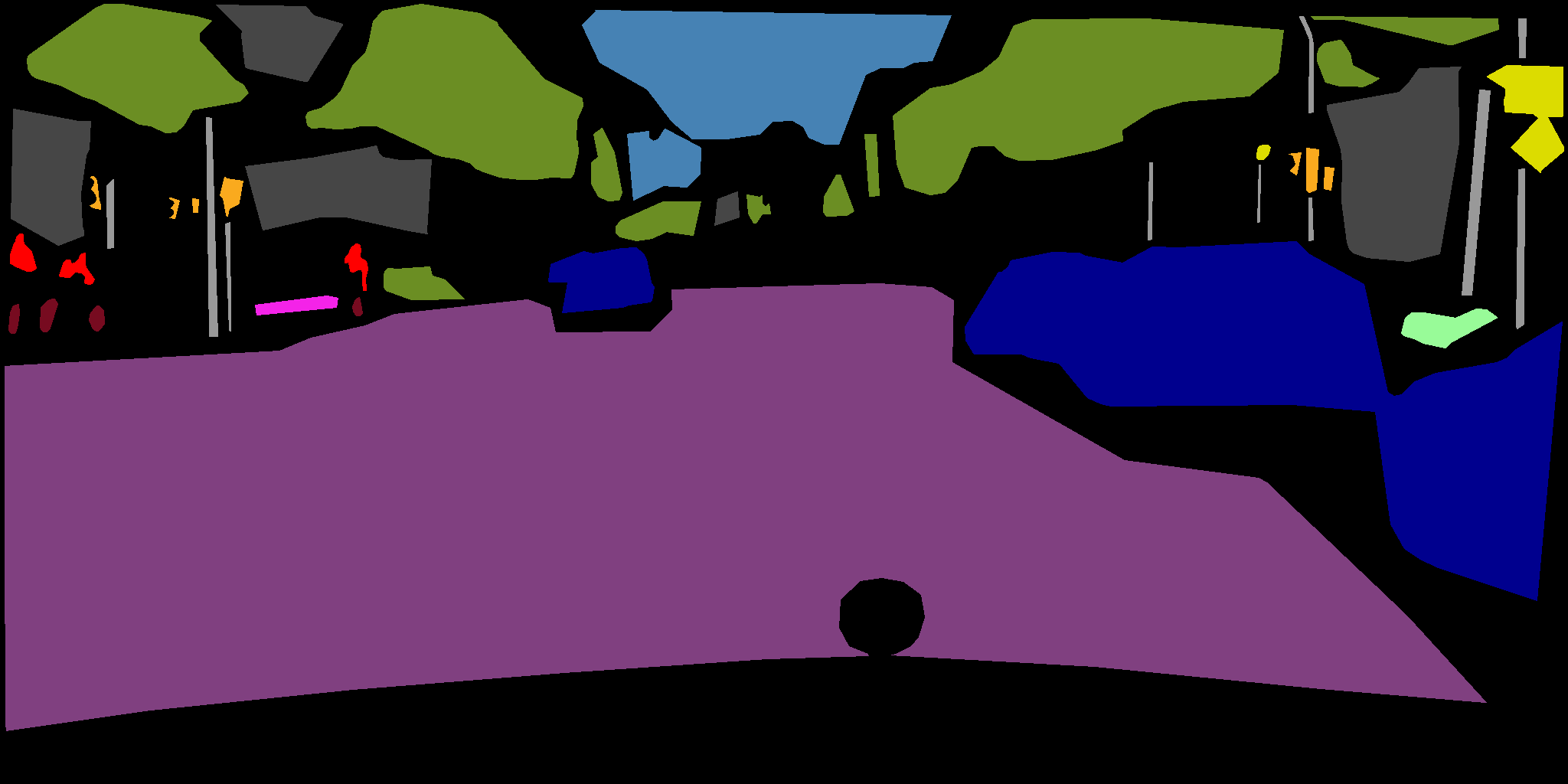} & \includegraphics[width=0.4\textwidth]{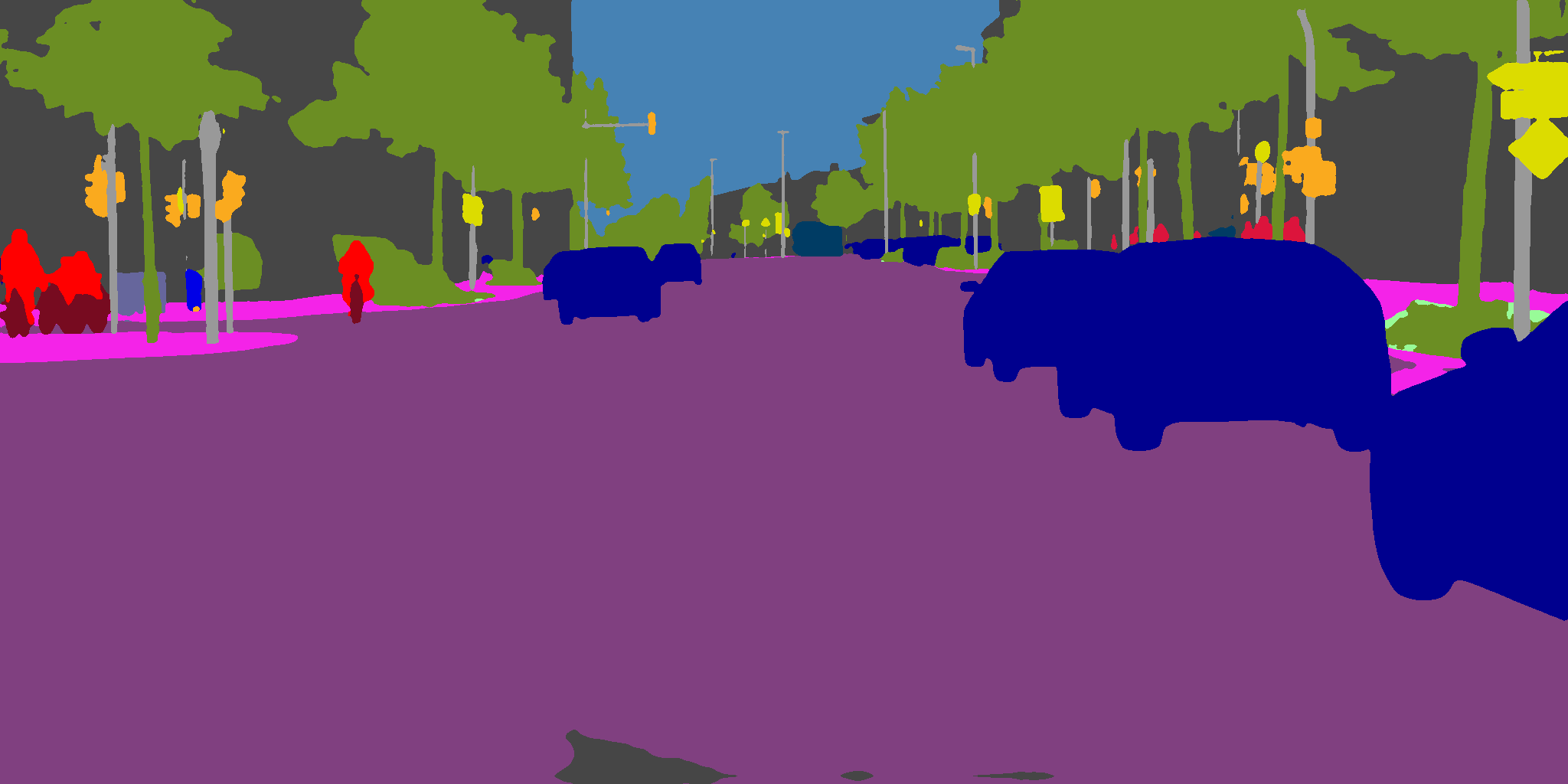}  \\
\includegraphics[width=0.4\textwidth]{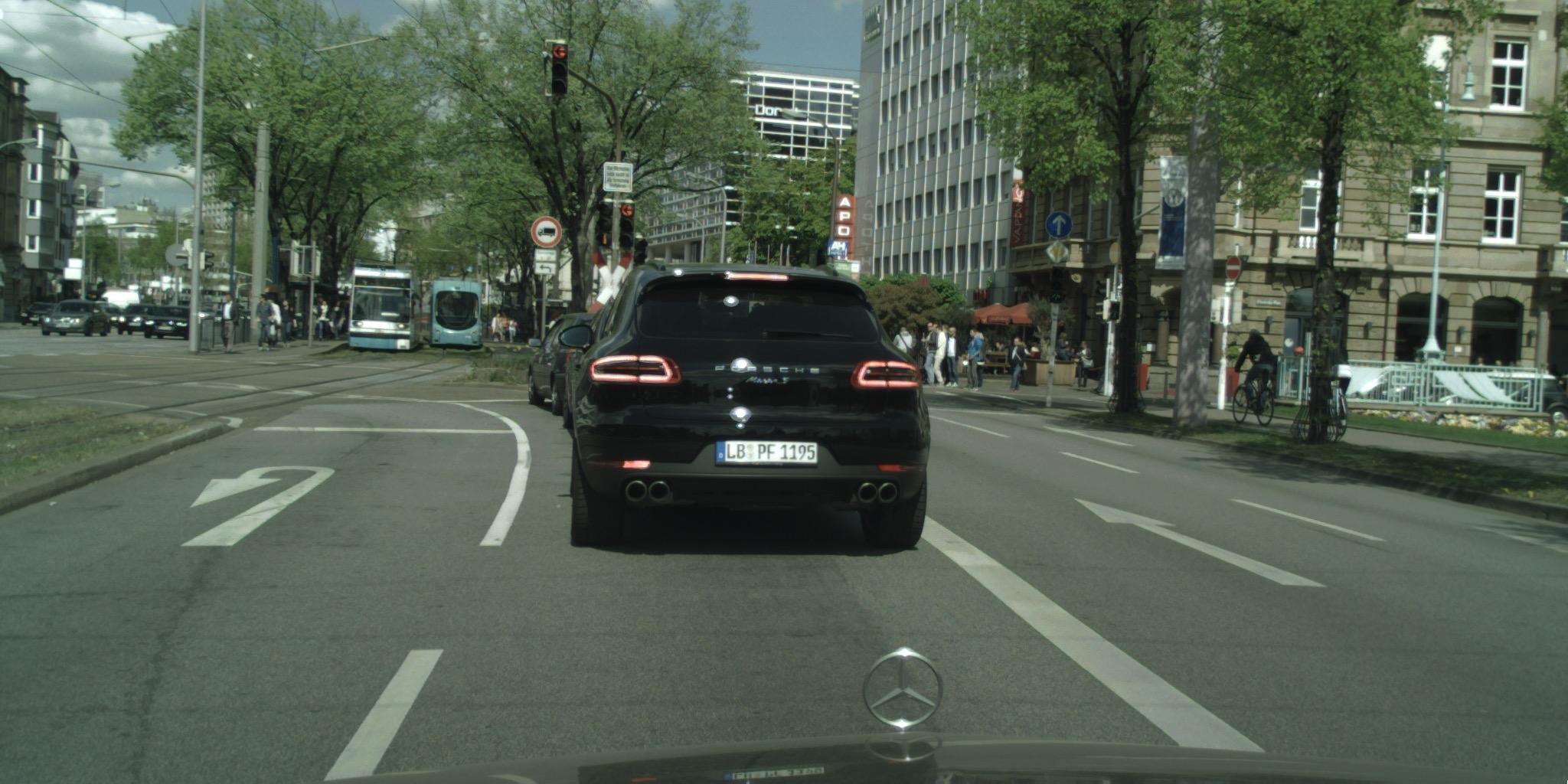} & \includegraphics[width=0.4\textwidth]{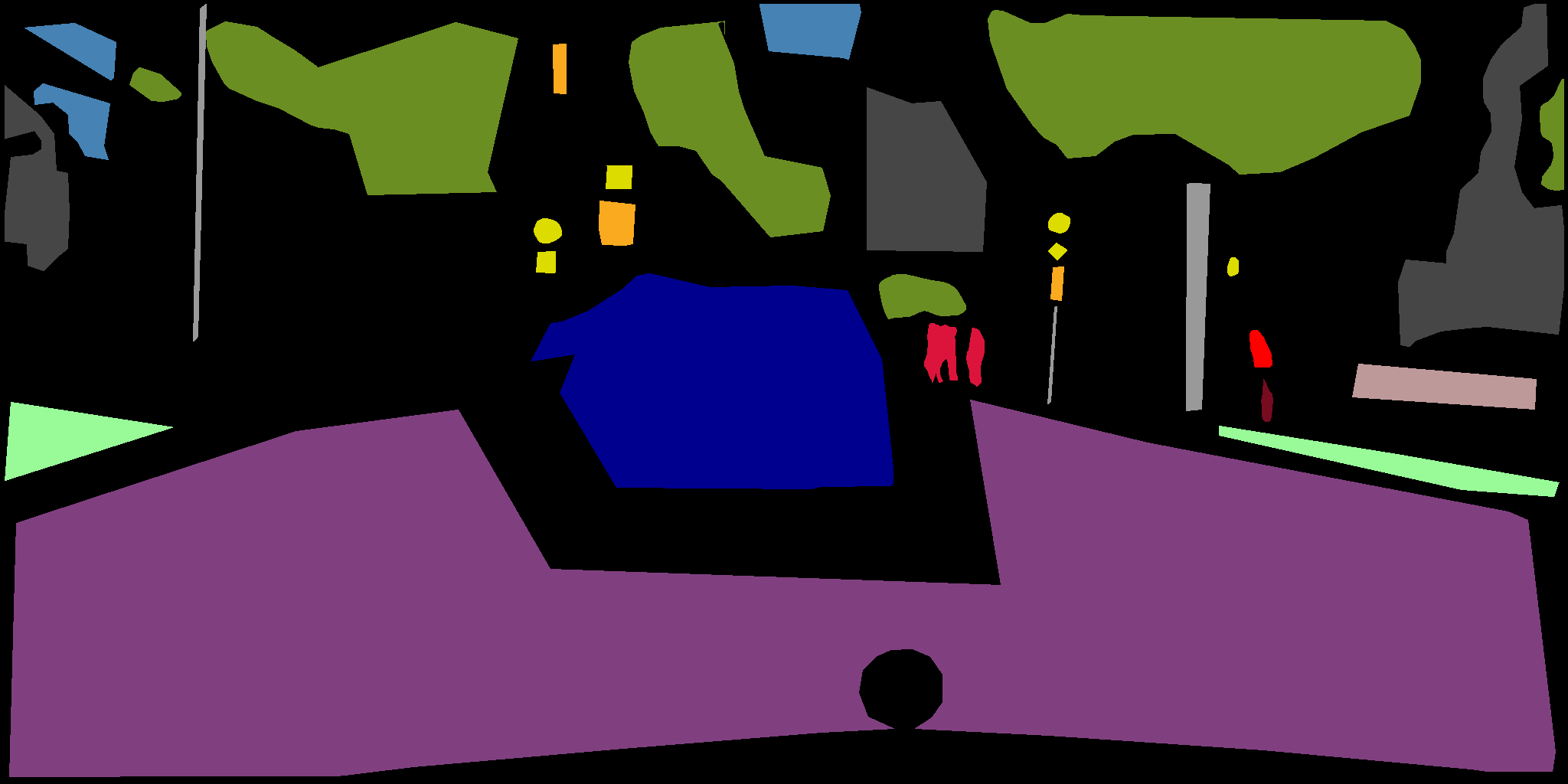} &\includegraphics[width=0.4\textwidth]{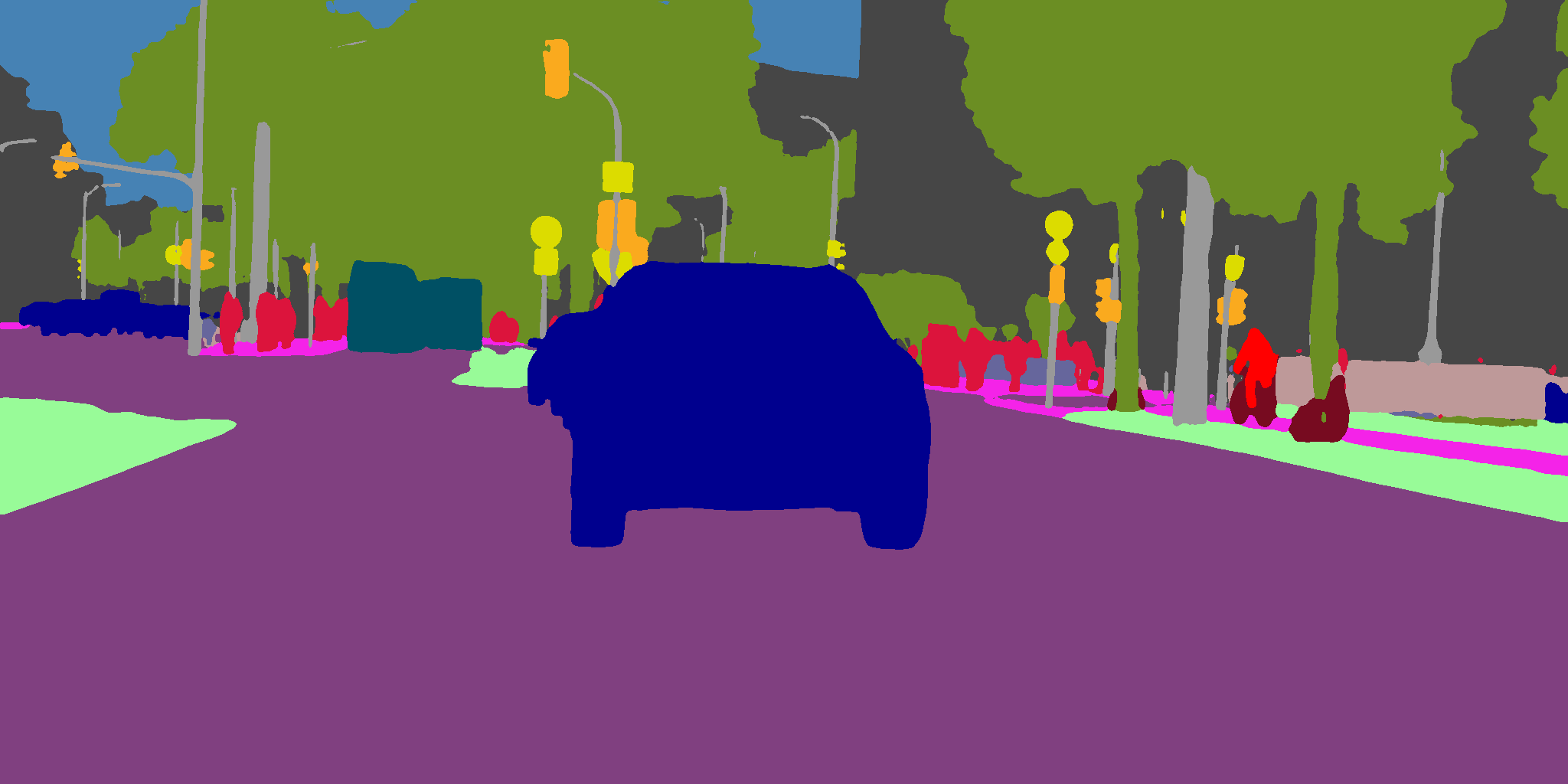} \\

\end{tabular}
}
\caption{Example of our auto-generated coarse image labels. Auto-generated coarse labels (right) provide finer detail of labelling than the original ground truth coarse labels (middle). This finer labelling improves the distribution of the labels since both small and large items are now represented, as opposed to primarily large items.}
\label{fig:teaser_comp}

\label{fig:fig4}
\end{figure}

Inspired by recent work on auto-labelling for image classification tasks ~\cite{xie2019selftraining} and ~\cite{tarvainen2017mean}, we adopt an auto-labelling strategy for Cityscapes to boost the effective dataset size and label quality. In Cityscapes, there 20,000 coarsely labelled images to go along with the 3,500 finely labelled images. The label quality of the coarse images is very modest and contains a large amount of unlabelled pixels, see Figure \ref{fig:fig4}. By using our auto-labelling approach, we can improve the label quality, which in turn helps the model IOU. 

A common technique for auto-labelling in image classification is to use \textbf{soft} or continuous labels, whereby a teacher network provides a target (soft) probability for each of N classes for every pixel of every image. A challenge of this approach is disk space and training speed: it costs roughly 3.2TB in disk space to store the labels: 20000 images * 2048 w * 1024 h * 19 classes * 4B = 3.2TB. Even if we chose to store such labels, reading such a volume of labels during training would likely slow training considerably.

Instead, we adopt a hard labelling strategy, whereby for a given pixel, we select the top class prediction of the teacher network. We threshold the label based on teacher network output probability. Teacher predictions that exceed the threshold become true labels, otherwise the pixel is labelled as~\textit{ignore} class. In practice we use a threshold of 0.9.

\section{Results}

\subsection{Implementation Protocol}
In this section, we describe our implementation protocol in detail.

\textbf{Training details} Our models are trained using Pytorch~\cite{NEURIPS2019_9015} on Nvidia DGX servers containing 8 GPUs per node with mixed precision, distributed data parallel training and synchronous batch normalization. We use Stochastic Gradient Descent (SGD) for our optimizer, with a batch size of $1$ per GPU, momentum $0.9$ and weight decay ${5e^{-4}}$ in training. We apply the “polynomial” learning rate policy ~\cite{liu2015parsenet}. We use RMI ~\cite{zhao2019rmi} as the the primary loss function under default settings, and we use cross-entropy for the auxiliary loss function. For Cityscapes, we use a poly exponent of $2.0$, an initial learning rate of $0.01$, and train for $175$ epochs across $2$ DGX nodes. For Mapillary, we use a poly exponent of $1.0$, an initial learning rate of $0.02$, and train for $200$ epochs across $4$ DGX nodes. As in ~\cite{semantic_cvpr19}, we use class uniform sampling in the data loader to equally sample from each class, which helps improve results when there is unequal data distribution.

\textbf{Data augmentation}: We employ gaussian blur, color augmentation, random horizontal flip and random scaling ($0.5$x - $2.0$x) on the input images to augment the dataset the training process. We use a crop size of $2048$x$1024$ for Cityscapes and $1856$x$1024$ for Mapillary.

\begin{table*}[]
\centering
\begin{tabular}{|cc|c|c|} \hline 
MS Attention      & Auto-labeling & IOU   & Gain \\ \hline \hline 
            &                &84.9  & \\
\checkmark  &                &85.4  &0.5 \\
            & \checkmark     &86.0  &1.1 \\
\checkmark  & \checkmark     &86.3  &1.4 \\
\hline 
\end{tabular}
\caption{Ablation study on Cityscapes \textit{validation} set. The baseline method uses HRNet-OCR as the architecture. \textbf{MS Attention} is our proposed multi-scale attention method. \textbf{Auto-labeling} indicates whether we are using automatically generated or ground truth coarse labels during training. A combination of both techniques yields the best results.}
\label{tab:table2}
\end{table*}

\subsubsection{Results on Cityscapes}
Cityscapes ~\cite{Cordts2016Cityscapes} is a large dataset that labels $19$ semantic classes across 5000 high resolution images. For Cityscapes, we use HRNet-OCR as the trunk along with our proposed multi-scale attention method. We use RMI as the loss for the main segmentation head but for the auxiliary segmentation head we use cross entropy because we found that using RMI loss led to reduced training accuracy deep into the training. Our best results are achieved by first pre-training on the larger Mapillary dataset, and then training on Cityscapes. For the Mapillary pre-training task, we do not train with attention. Our state-of-the-art recipe on Cityscapes was achieved using train + val images in addition to the auto-labelled coarse images. At 50\% probability we sample from the train + val set, else we sample from the auto-labelled pool of images. At inference time, we use scales = \{$0.5$, $1.0$, $2.0$\} and image flipping.

We conduct ablation studies on Cityscapes validation set as shown in Table \ref{tab:table2}. Multi-scale attention yields $0.5$\% IOU over the baseline HRNet-OCR architecture with average pooling. Auto-labelling provides a boost of 1.1\% IOU over the baseline. Combining both techniques together results in a total gain of 1.4\% IOU.

\begin{figure*}
\centering

\begin{tabular}{ccc}
 \\
 Input images & Ground truth & Our network prediction \\
 \includegraphics[width=0.3\textwidth]{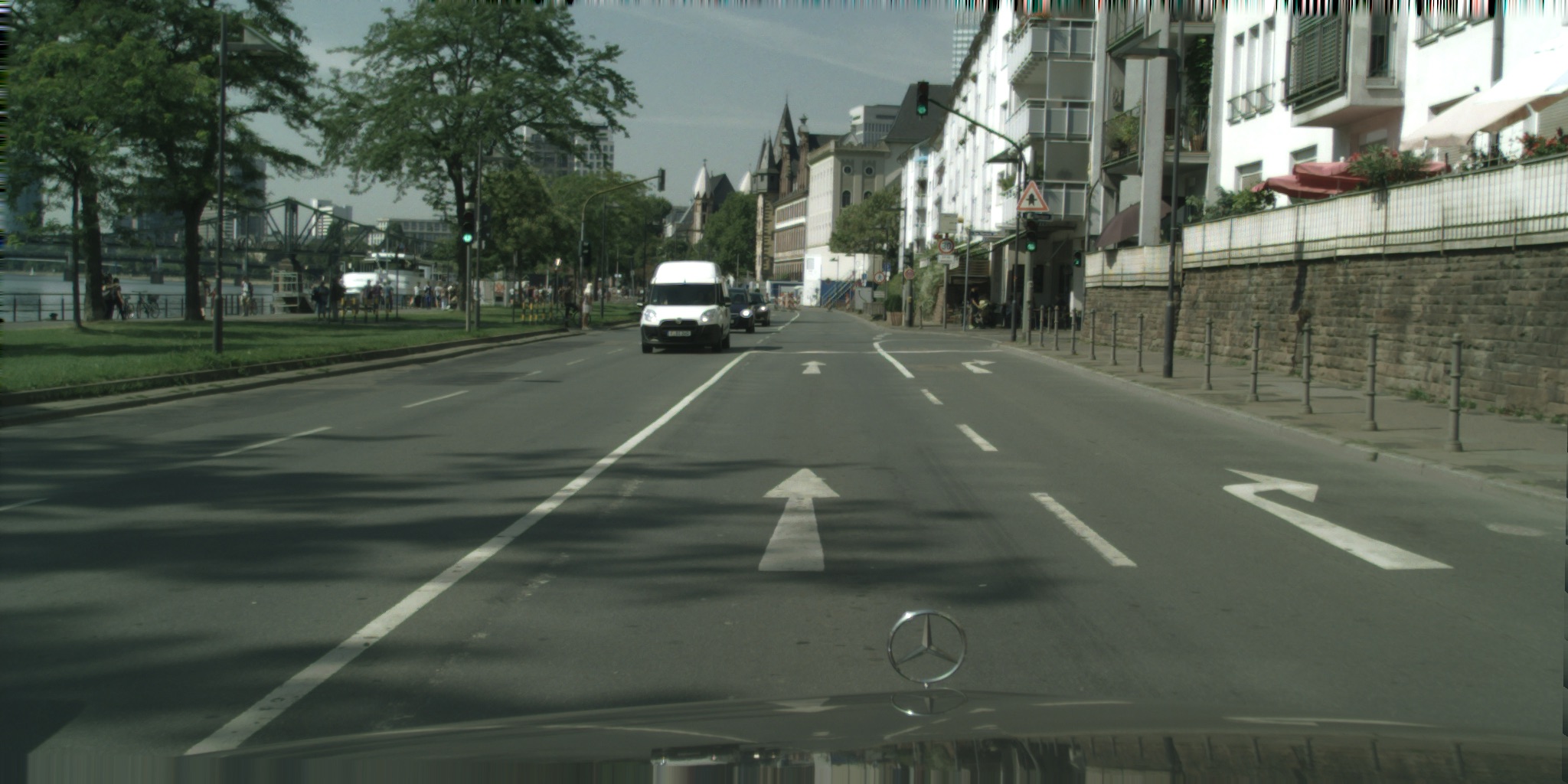} & \includegraphics[width=0.3\textwidth]{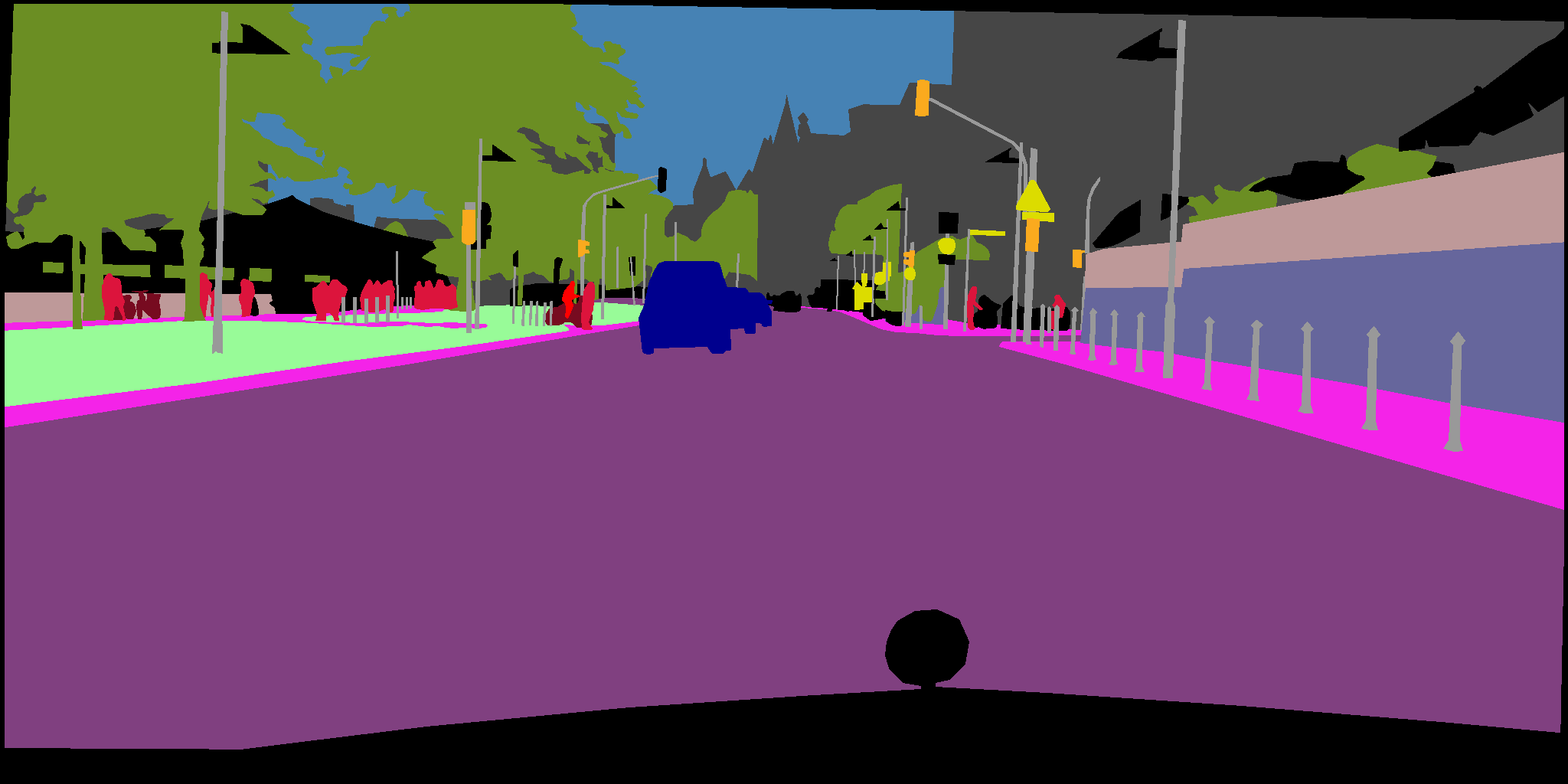} & \includegraphics[width=0.3\textwidth]{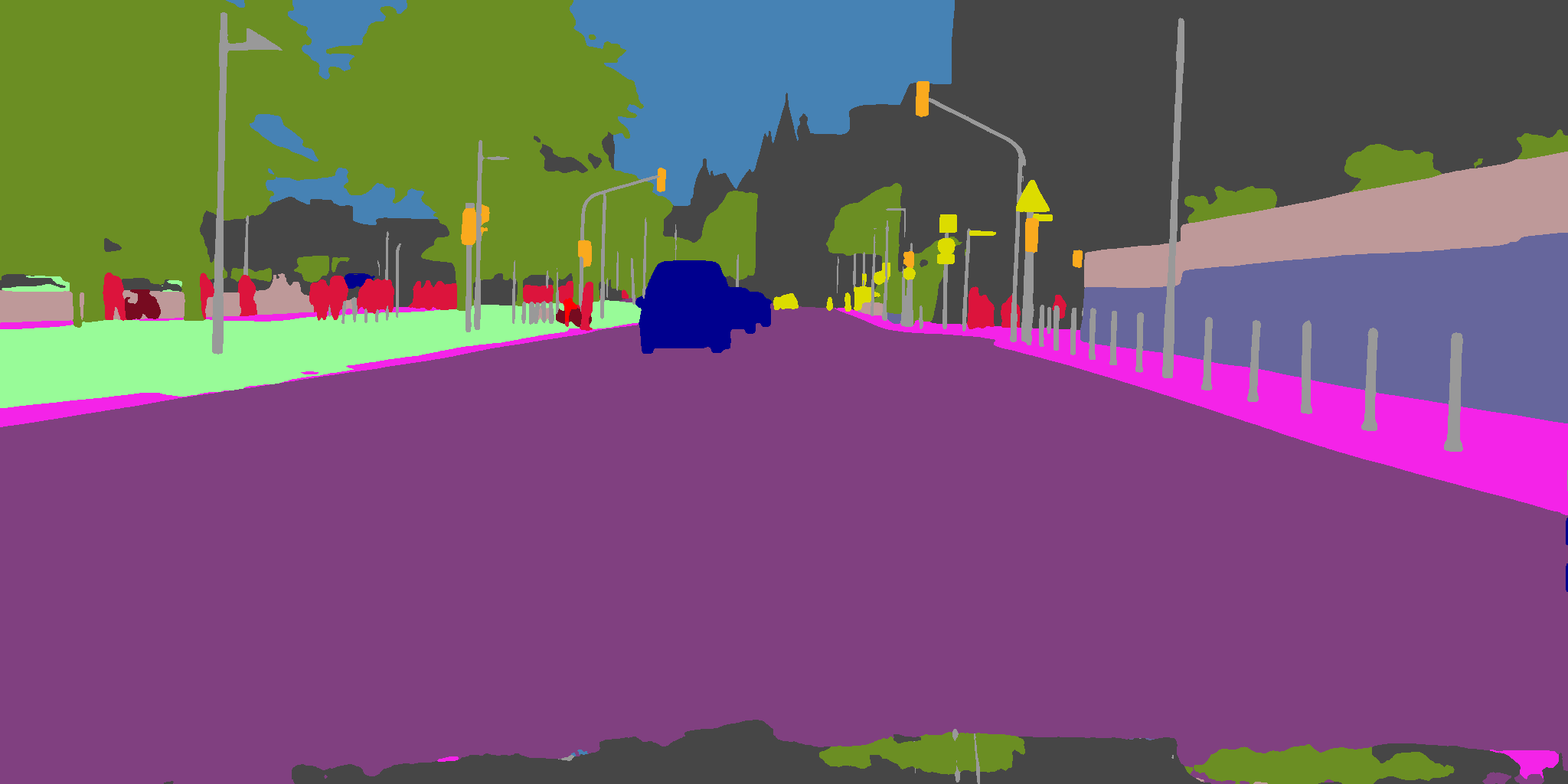} \\
\includegraphics[width=0.3\textwidth]{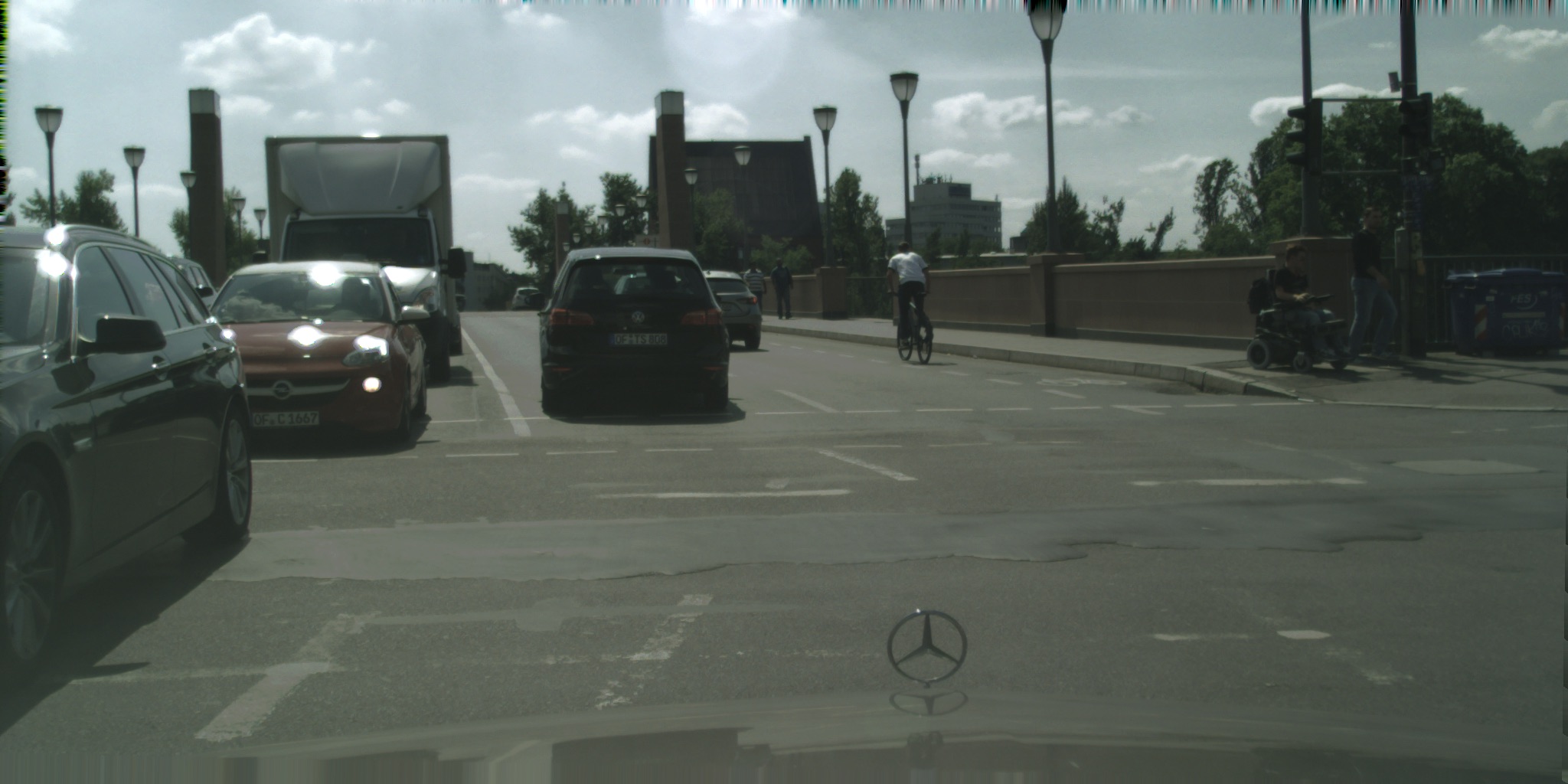} & \includegraphics[width=0.3\textwidth]{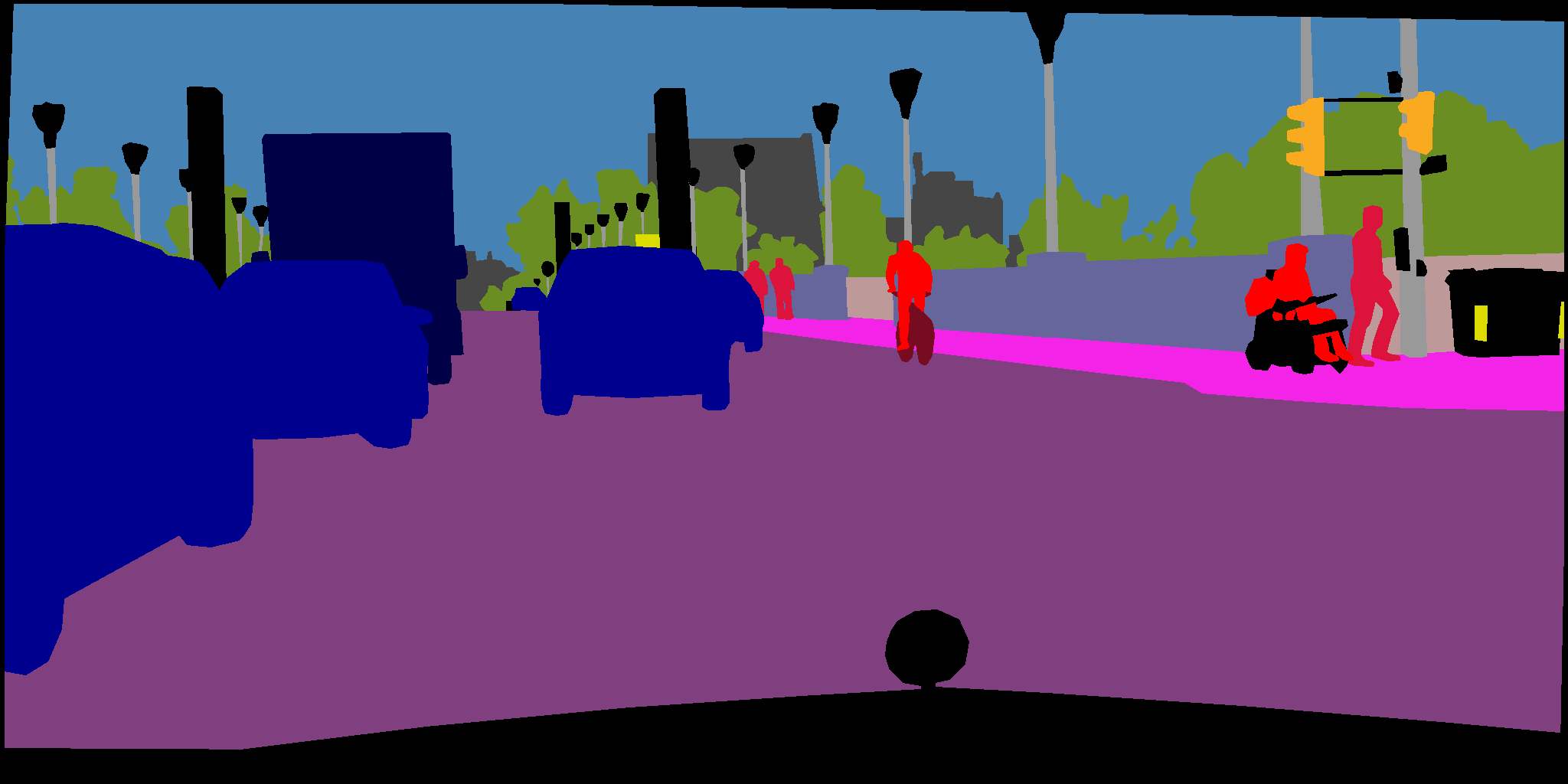} & \includegraphics[width=0.3\textwidth]{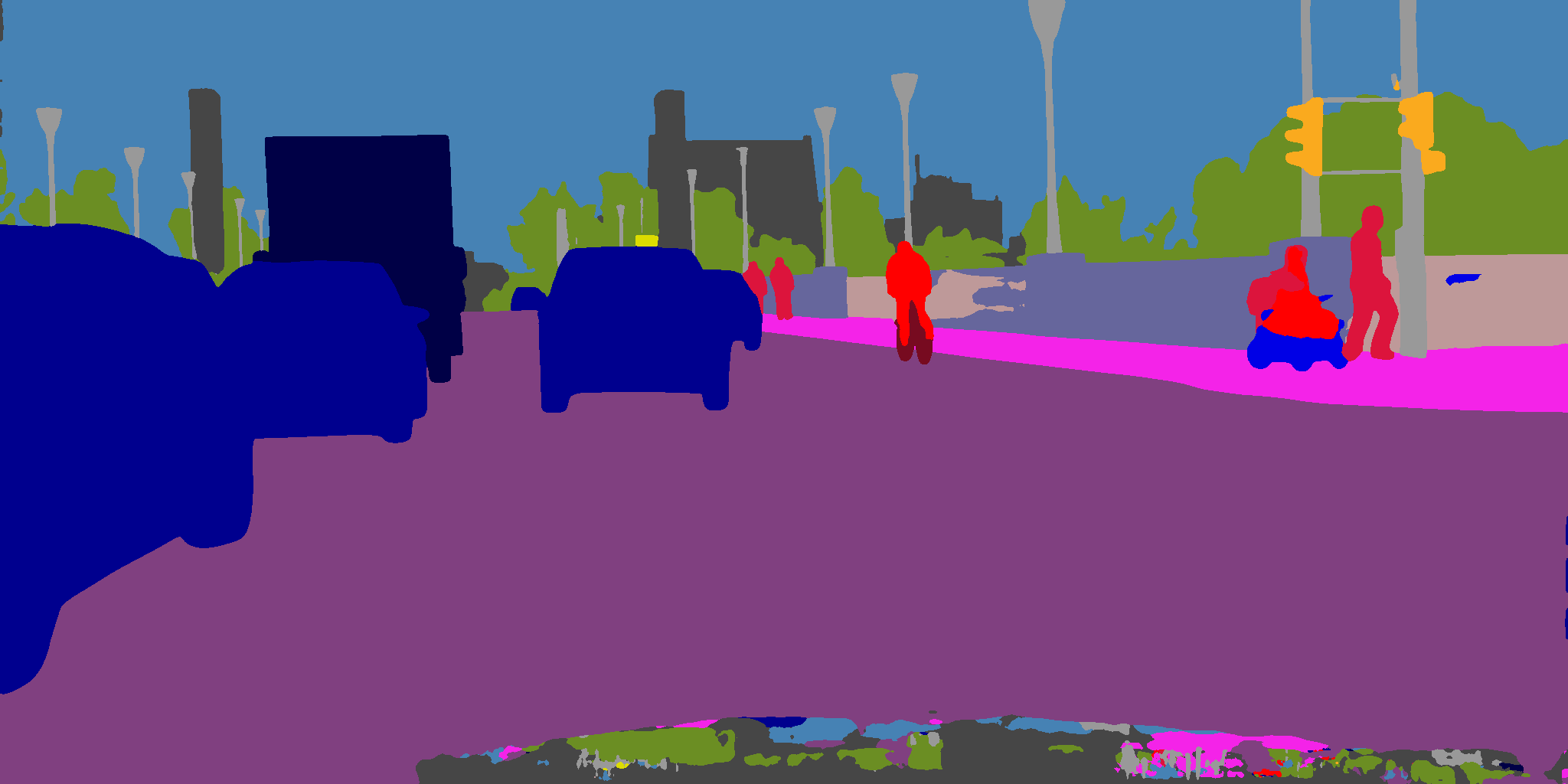} \\
\includegraphics[width=0.3\textwidth]{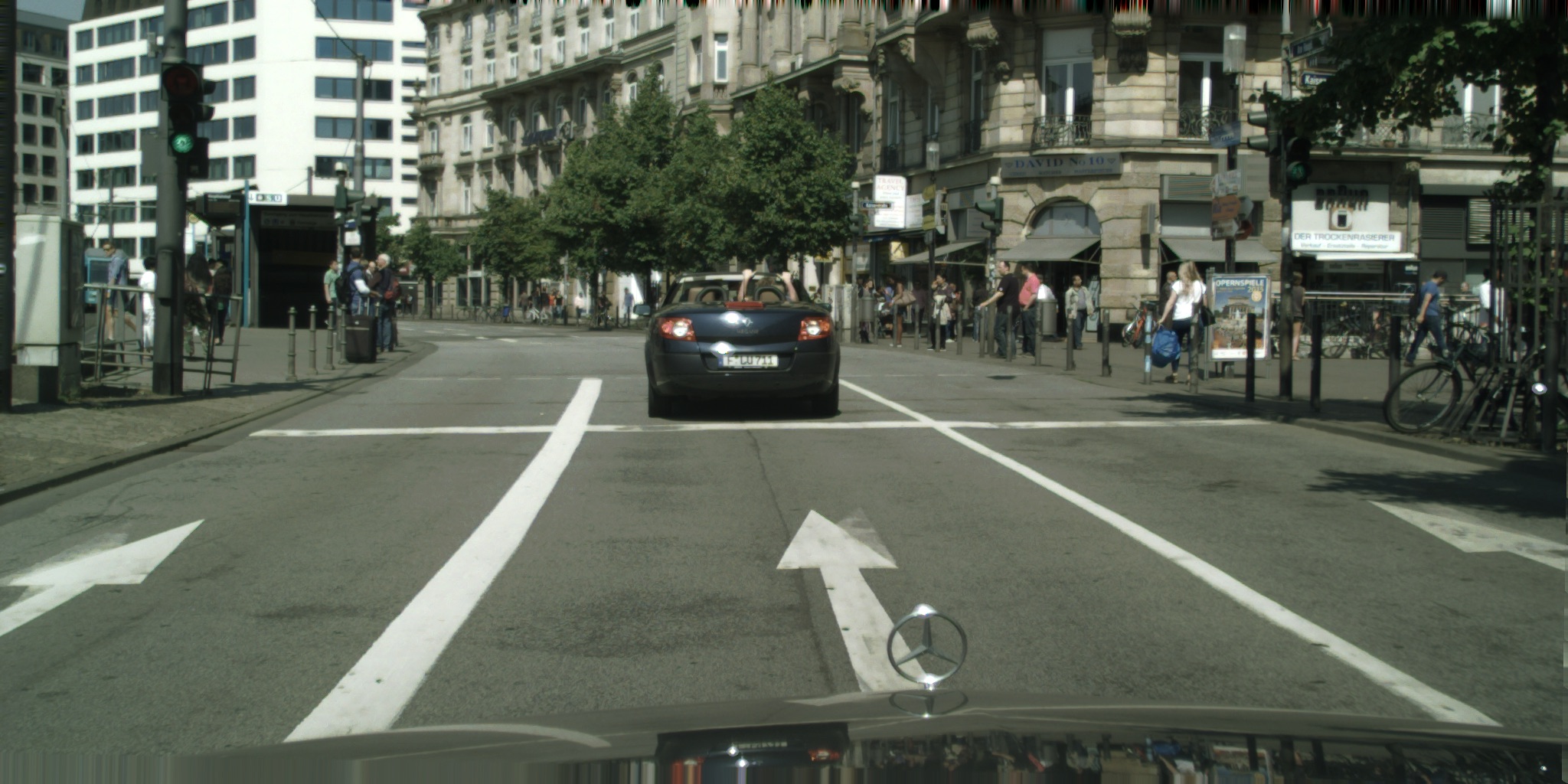} & \includegraphics[width=0.3\textwidth]{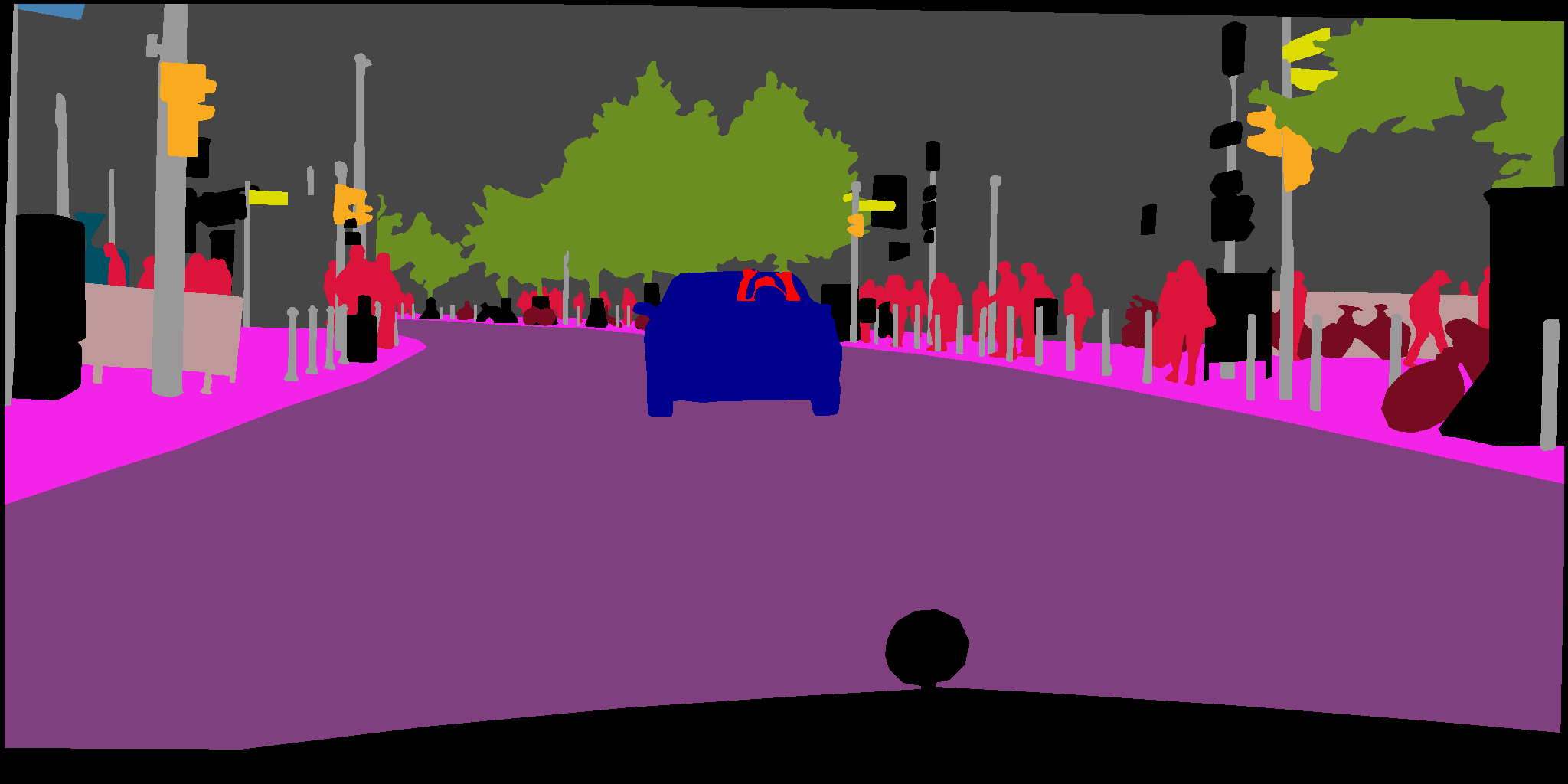} & \includegraphics[width=0.3\textwidth]{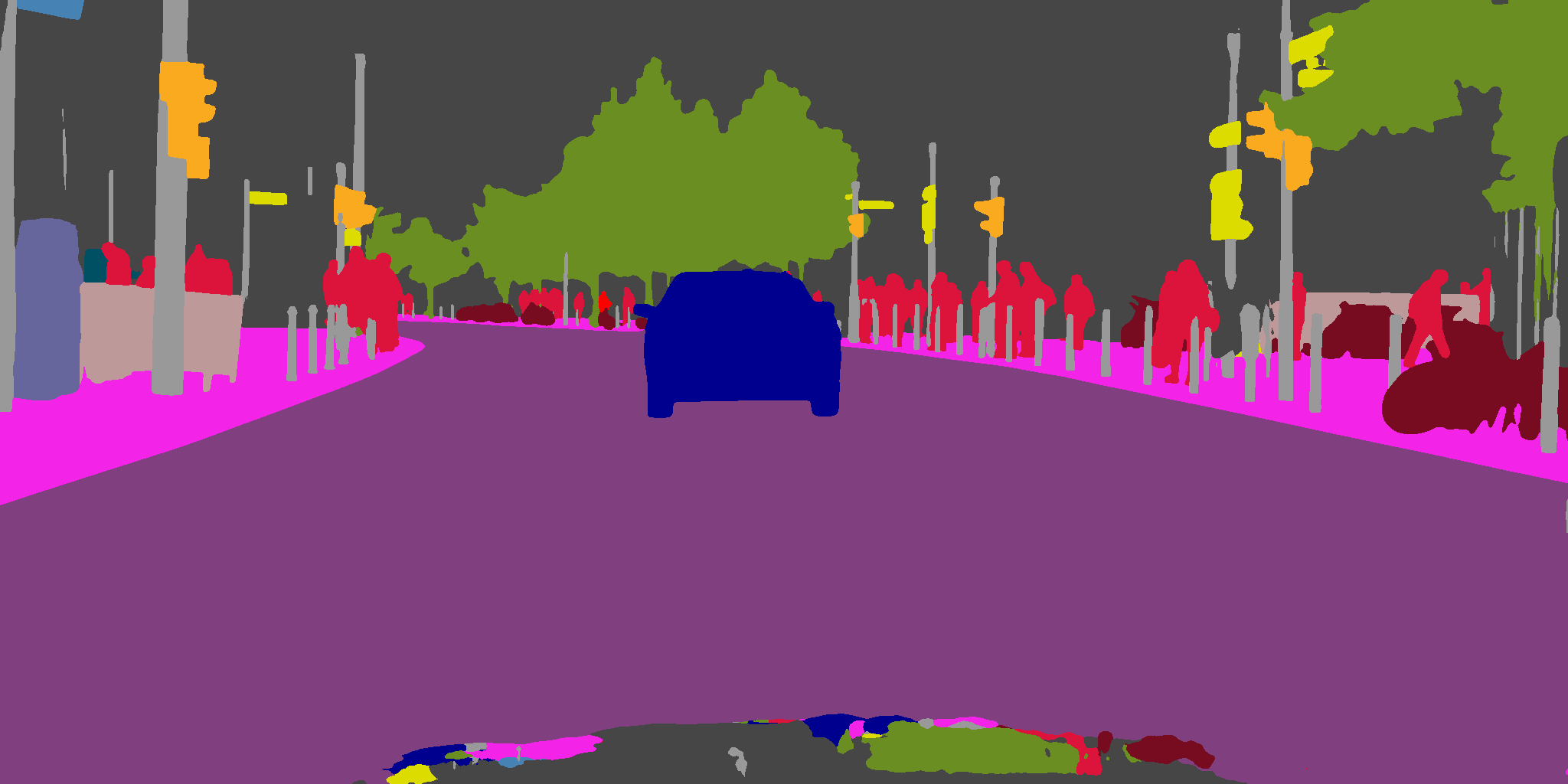} \\
\includegraphics[width=0.3\textwidth]{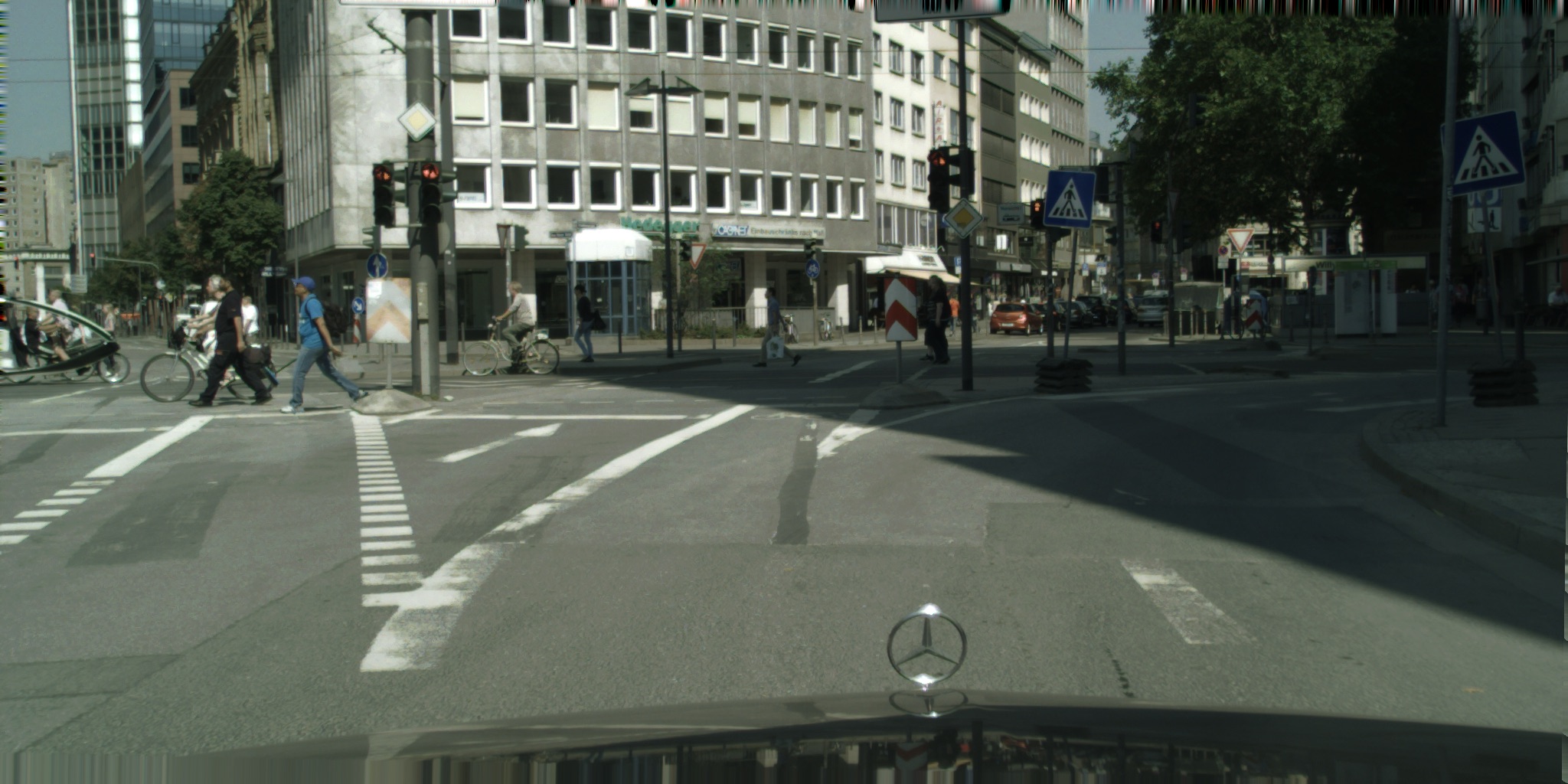} & \includegraphics[width=0.3\textwidth]{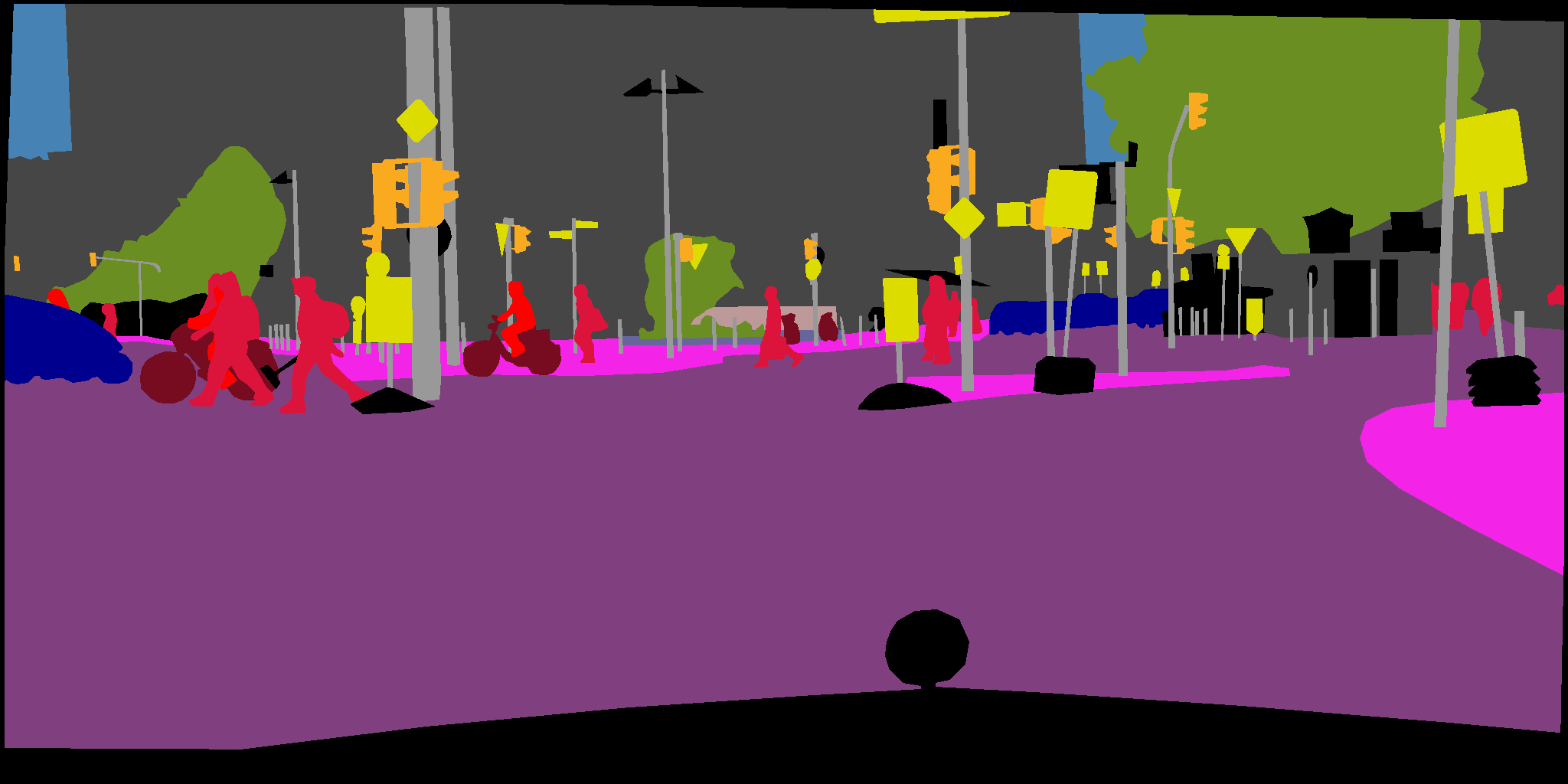} & \includegraphics[width=0.3\textwidth]{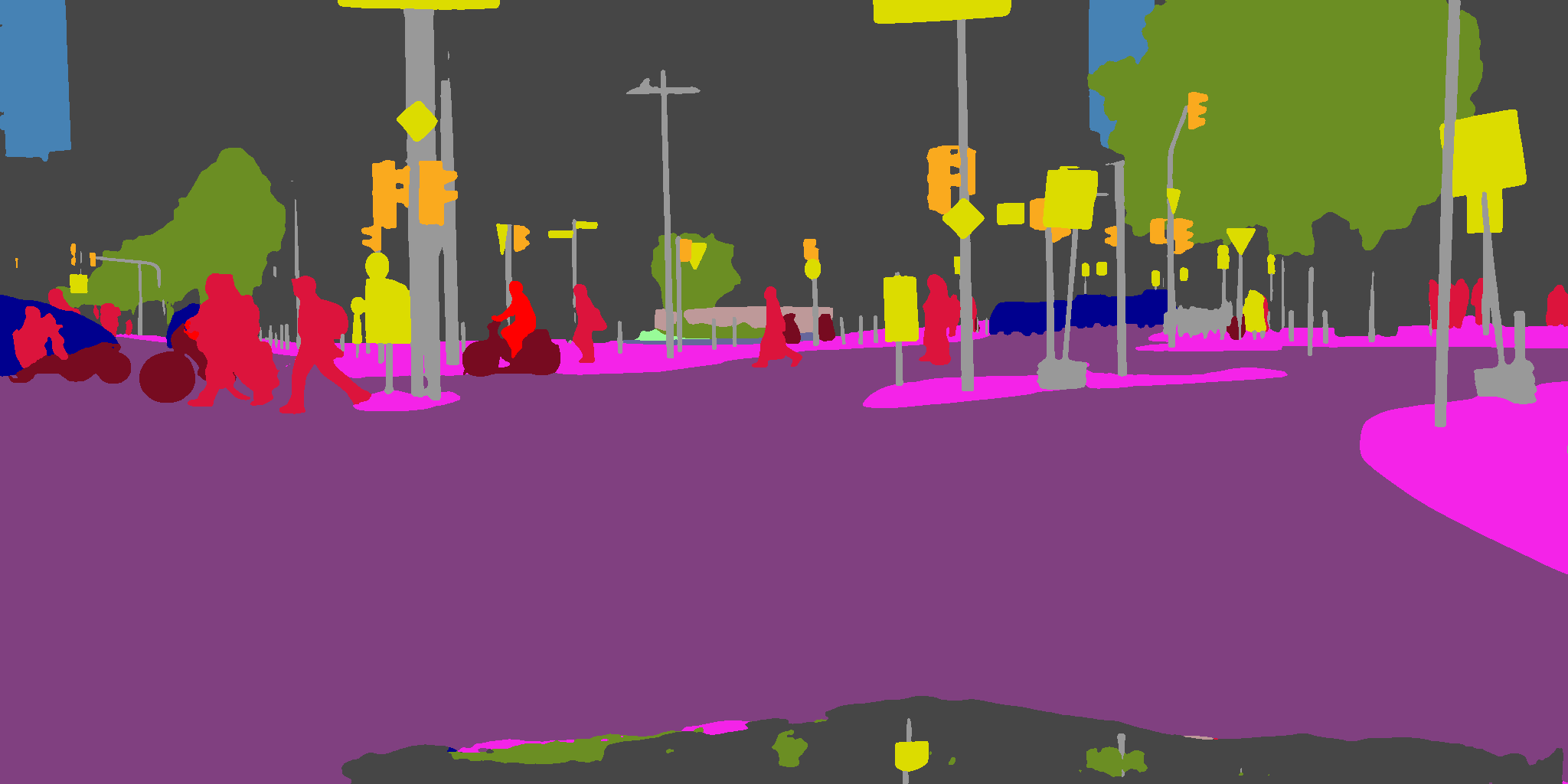} \\
\includegraphics[width=0.3\textwidth]{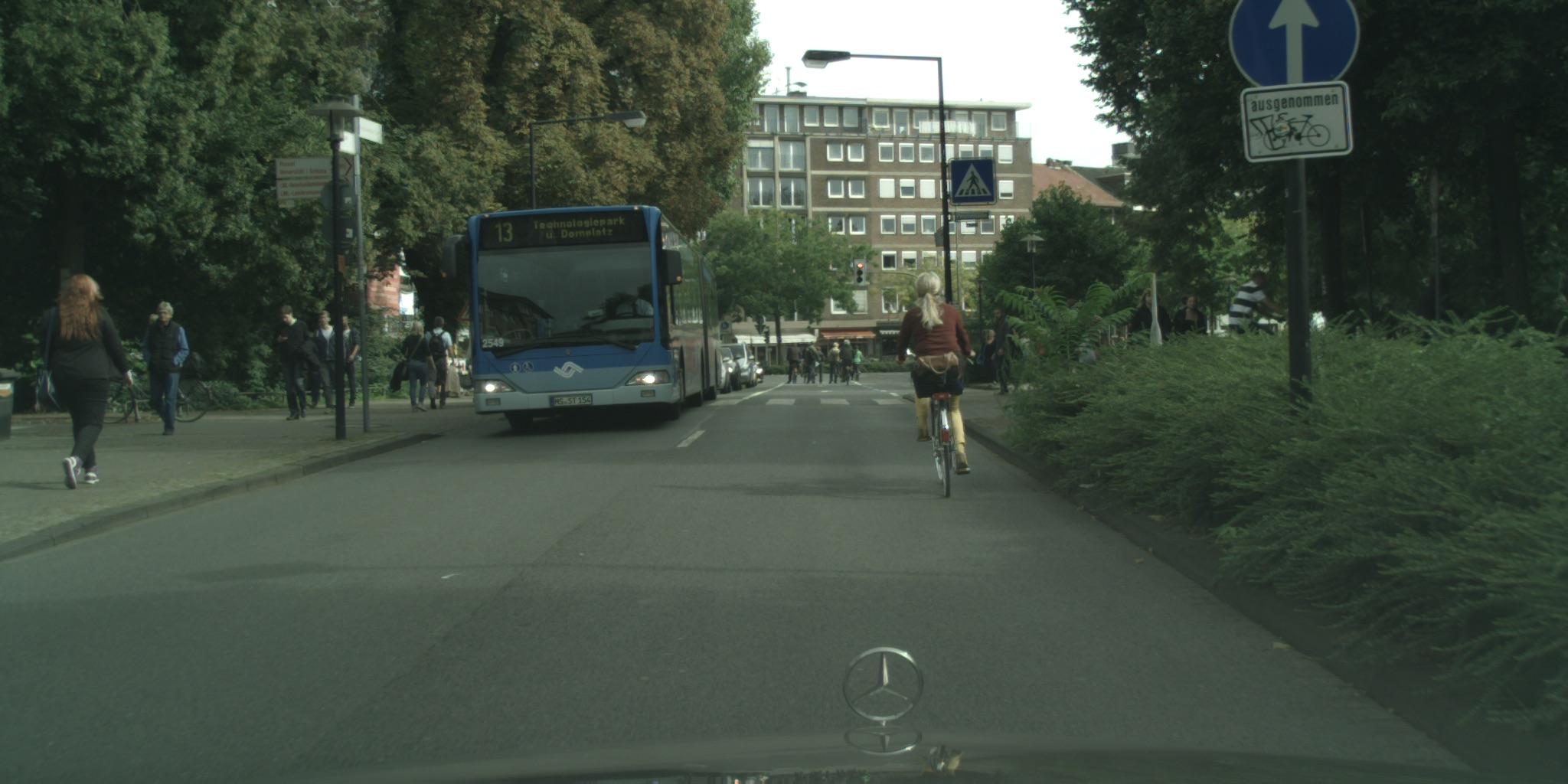} & \includegraphics[width=0.3\textwidth]{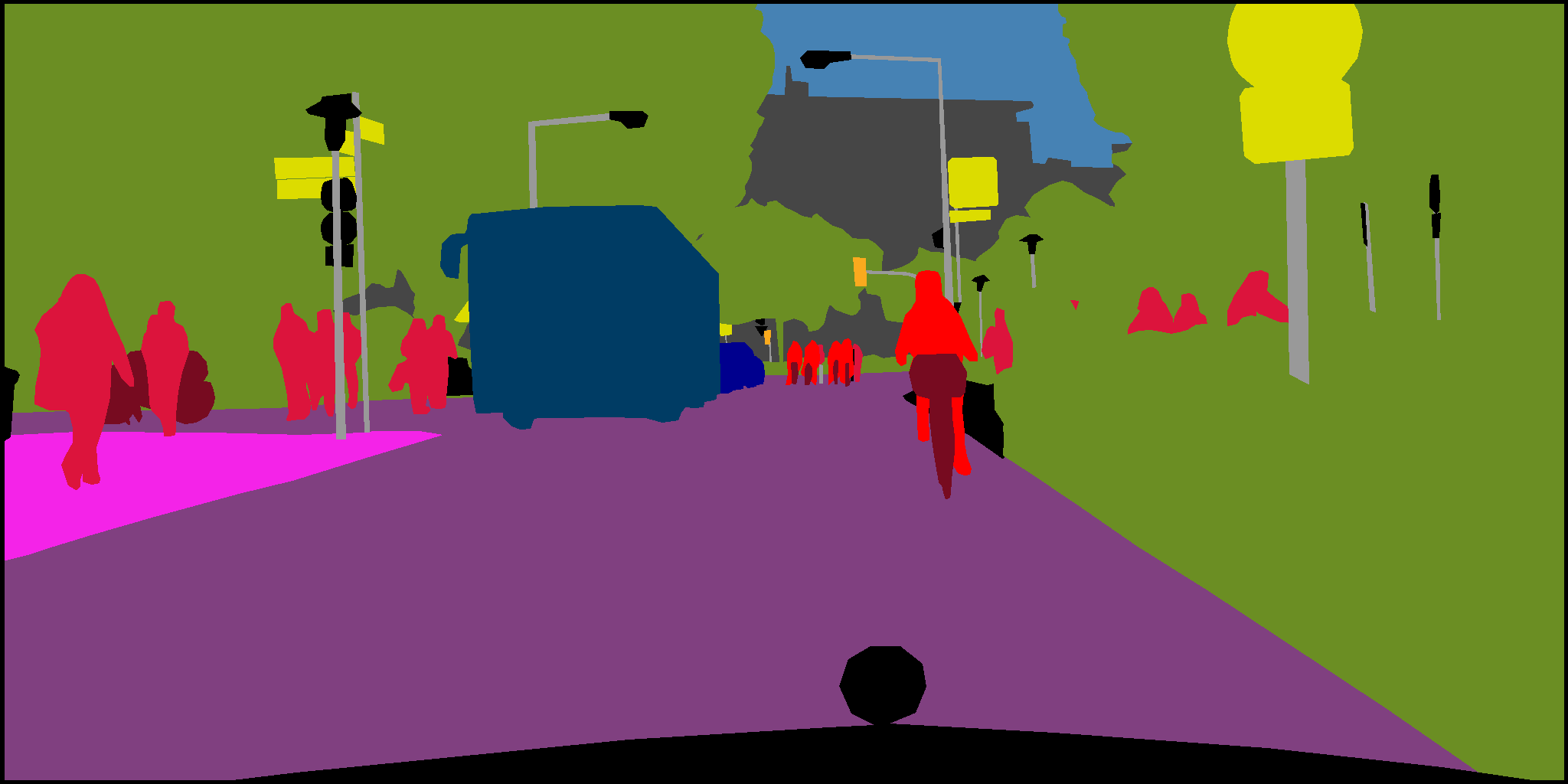} & \includegraphics[width=0.3\textwidth]{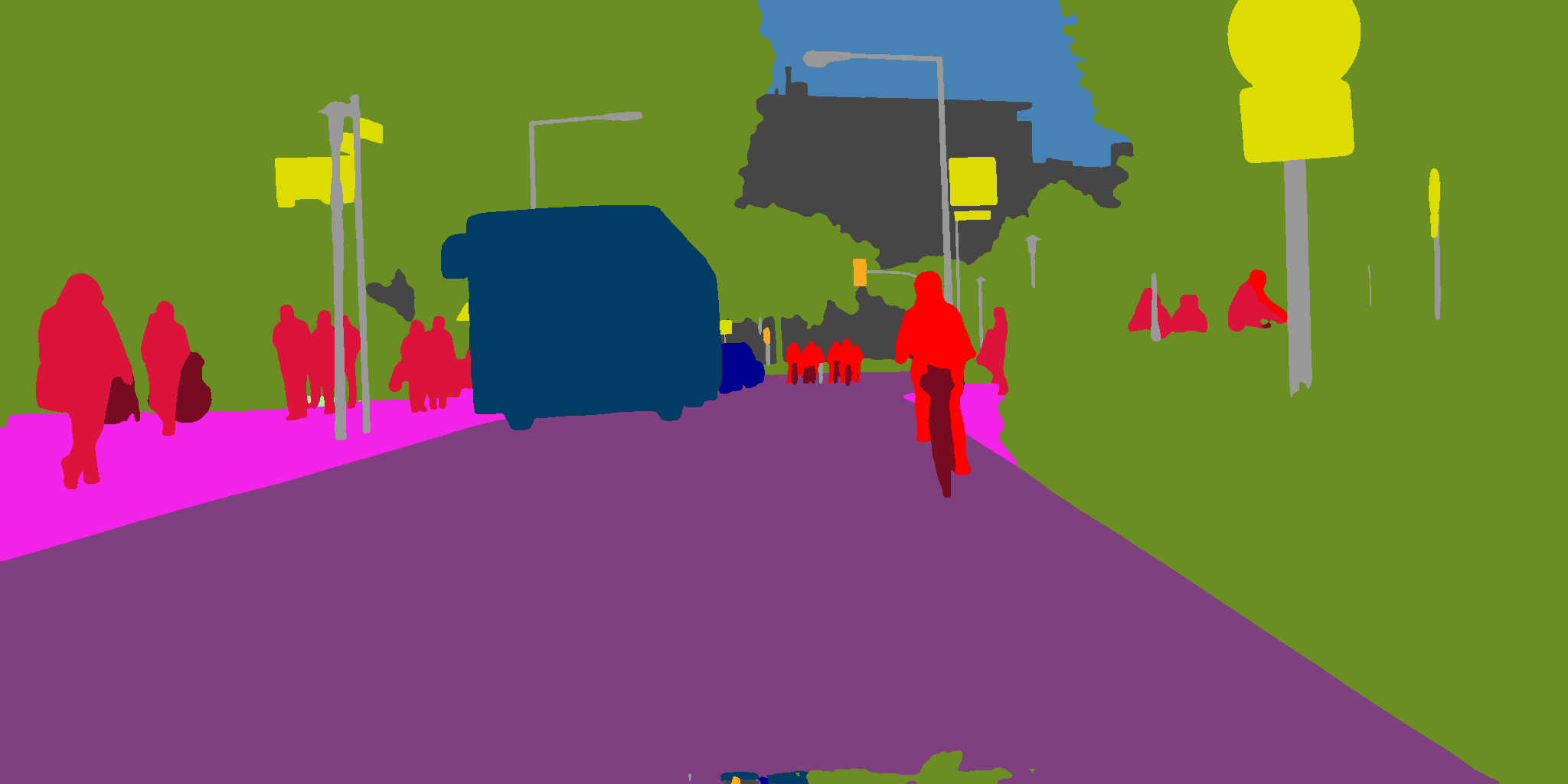} \\
\includegraphics[width=0.3\textwidth]{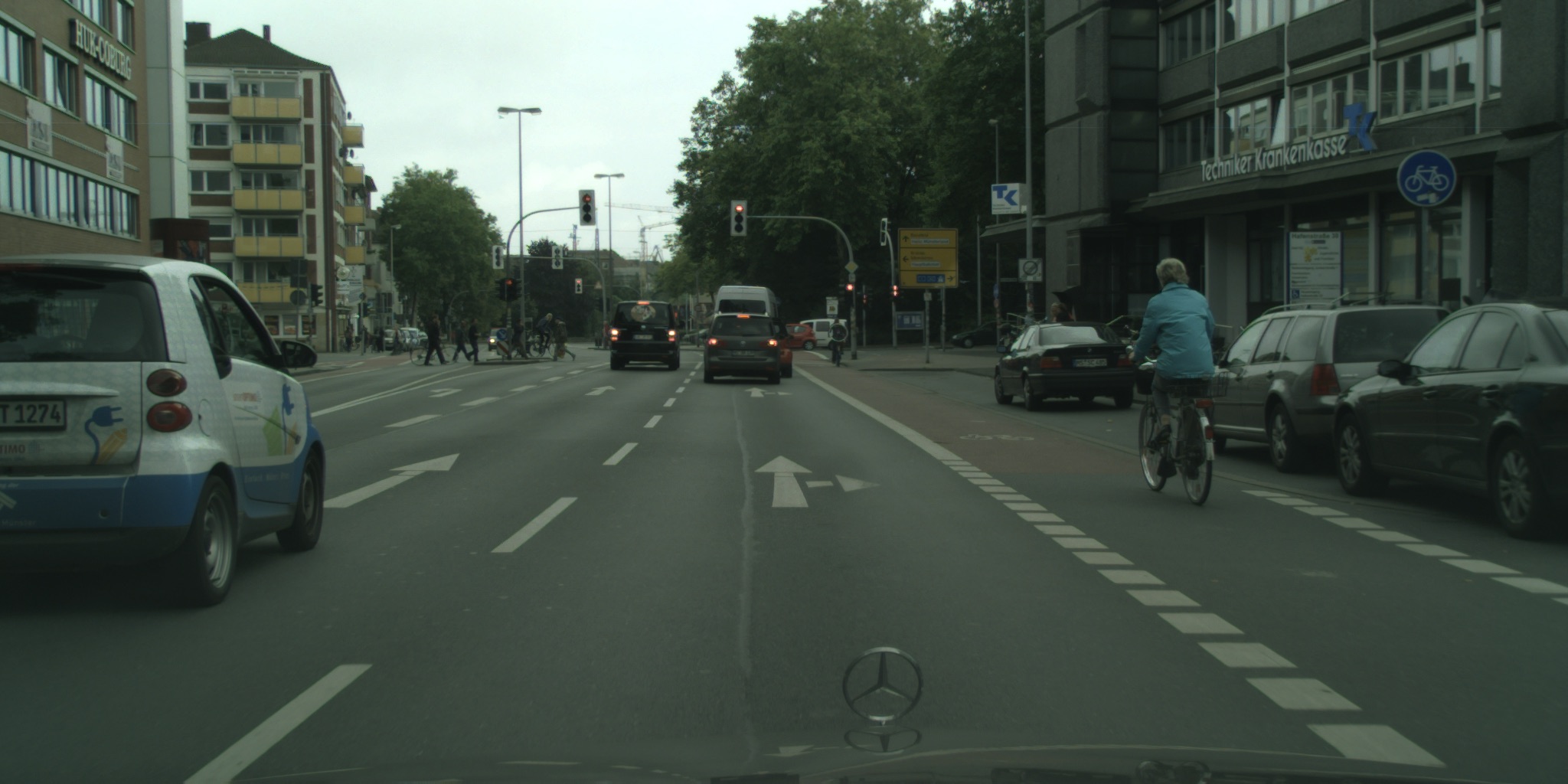} & \includegraphics[width=0.3\textwidth]{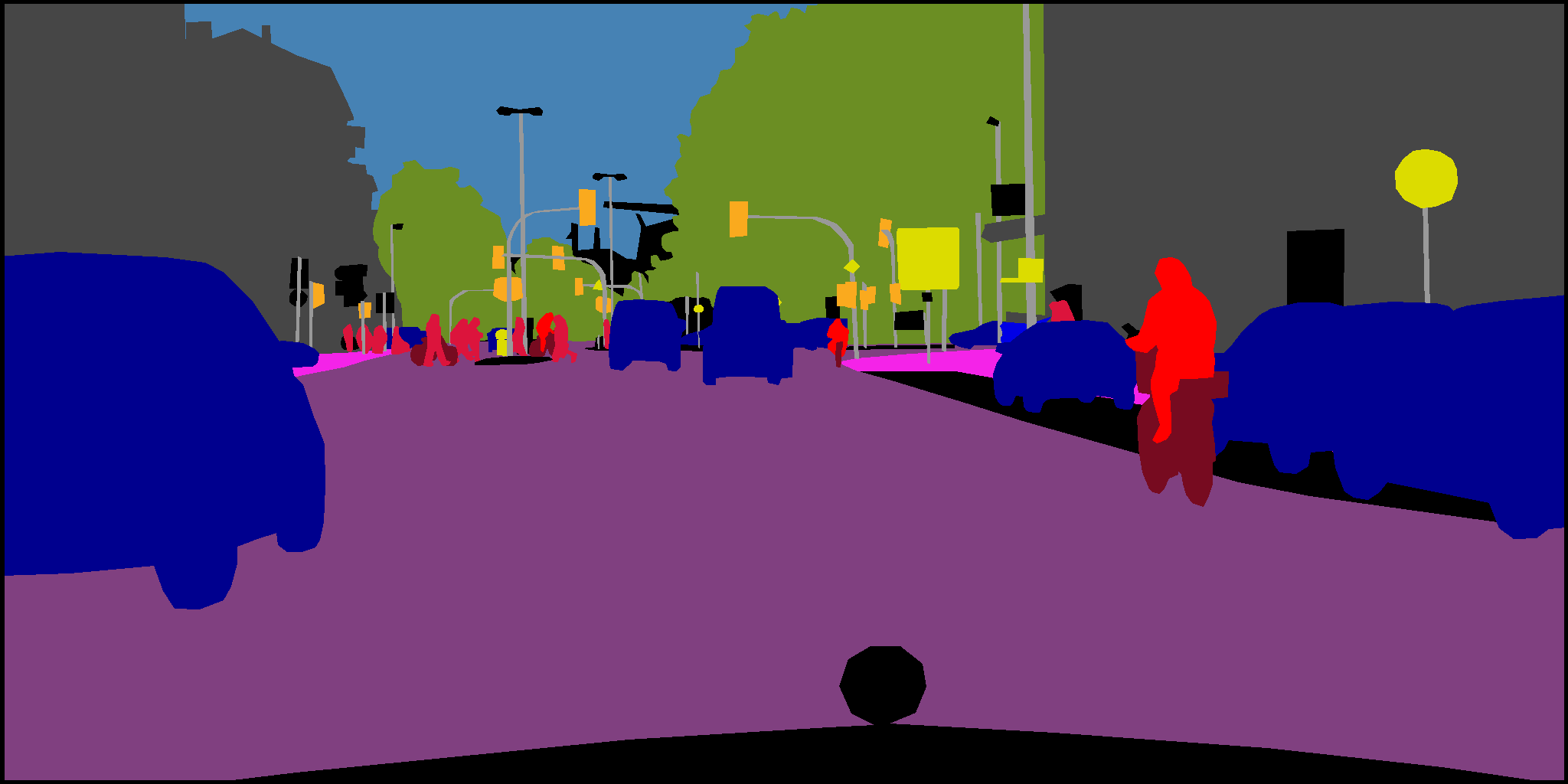} & \includegraphics[width=0.3\textwidth]{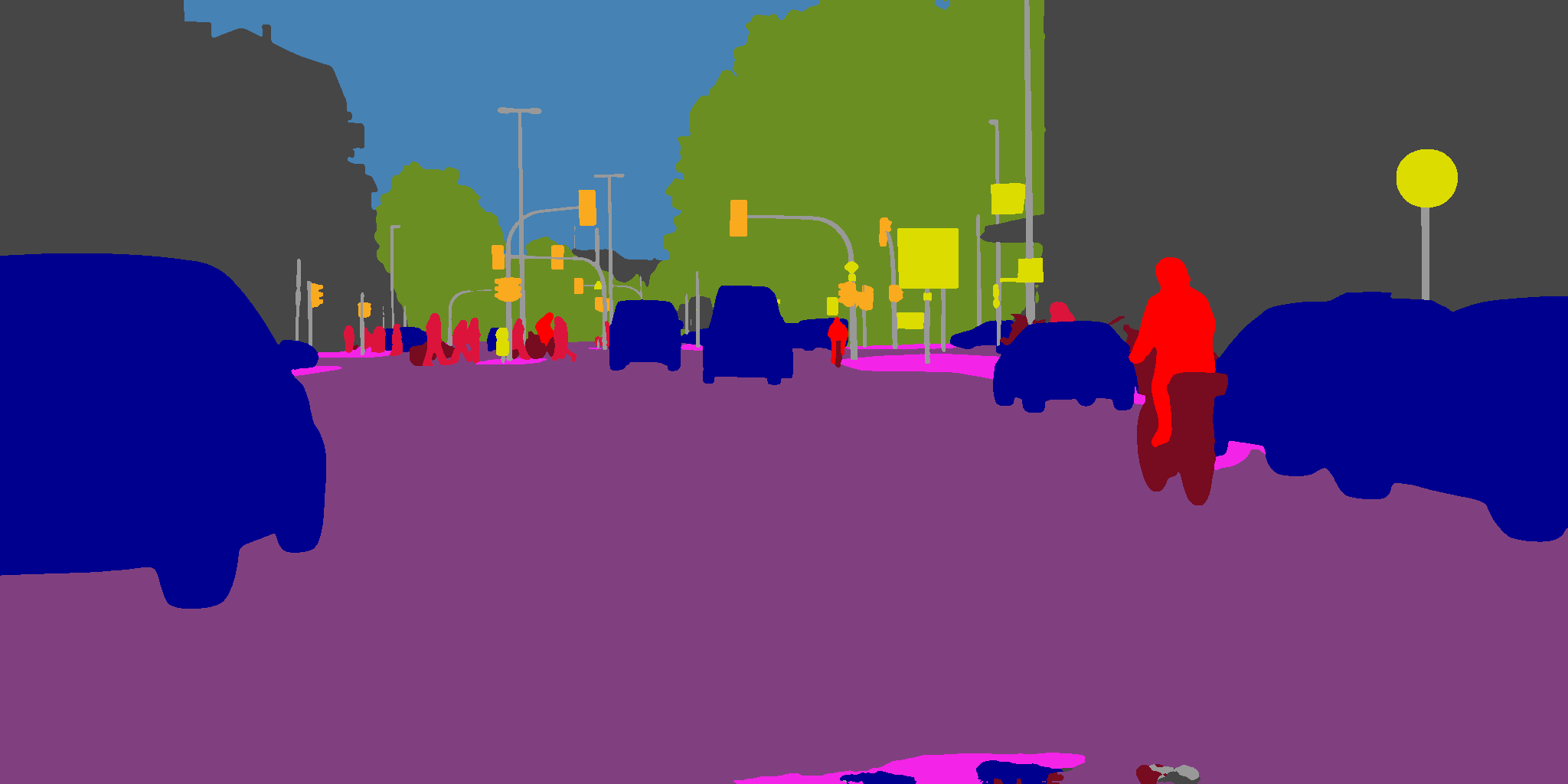}

\end{tabular}
\caption{Qualitative Results. From left to right: input, ground truth, our method on Cityscapes.}
\label{fig:cityscapes_results}
\vspace{-0.5cm}
\end{figure*}

Finally, in Table \ref{tab:table4} we show results of our method as compared to other top-performing methods in the Cityscapes test set. Our method achieves a score of $85.1$, which is the best reported Cityscapes test score of all methods, beating the best previous score by $0.6$ IOU. In addition, our method has the top per-class scores in all but three classes. Some results are visualized in Figure~\ref{fig:cityscapes_results}.

\begin{table*}[t]
	\begin{center}
		\resizebox{1.0\columnwidth}{!}
		{%
			\begin{tabular}{  c | ccccccccccccccccccc| c }
				\toprule
				Method	&  road  & swalk & build. & wall & fence & pole & tlight & tsign & veg. & terrain & sky & person & rider & car & truck & bus & train & mcycle &  bicycle & mIoU   \\  \hline  \hline
			VPLR~\cite{semantic_cvpr19}  & 98.8 & 87.8 & 94.2 & 64.1 & 65.0 & 72.4 & 79.0 & 82.8 & 94.2 & 74.0 & 96.1 & 88.2 & 75.4 & 96.5 & 78.8 & 94.0 & 91.6 & 73.7 & 79.0 & 83.5 \\
HRNet-OCR ASPP~\cite{yuan2019objectcontextual}   & 98.8 & 88.3 & 94.3 & 66.9 & 66.7 & 73.3 & 80.2 & 83.0 & 94.2 & 74.1 & 96.0 & 88.5 & 75.8 & 96.5 & 78.5 & 91.8 & 90.1 & 73.4 & 79.3 & 83.7 \\
Panoptic Deeplab~\cite{cheng2019panopticdeeplab} & 98.8 & 88.1 & 94.5 & 68.1 & 68.1 & 74.5 & 80.5 & 83.5 & 94.2 & 74.4 & 96.1 & 89.2 & 77.1 & 96.5 & 78.9 & 91.8 & 89.1 & 76.4 & 79.3 & 84.2 \\
iFLYTEK-CV  & 98.8 & 88.4  & 94.4  & 68.9 & 66.8 & 73.0 & 79.7 & 83.3 & 94.3  & 74.3 & 96.0 & 88.8 & 76.3 & 96.6 & \textbf{84.0} & \textbf{94.3} & \textbf{91.7} & 74.7 & 79.3  & 84.4\\
SegFix~\cite{yuan2020segfix} & 98.8 & 88.3 & 94.3 & 67.9 & 67.8 & 73.5& 80.6 & 83.9& 94.3 & 74.4 & 96.0 & 89.2 & 75.8& 96.8 & 83.6 & 94.1 & 91.2& 74.0 & 80.0 & 84.5 \\

Ours    & \textbf{99.0} & \textbf{89.2} & \textbf{94.9} & \textbf{71.6} & \textbf{69.1} & \textbf{75.8} & \textbf{82.0} & \textbf{85.2} & \textbf{94.5} & \textbf{75.0} & \textbf{96.3} & \textbf{90.0} & \textbf{79.4} & \textbf{96.9} & 79.8 & 94.0 & 85.8 & \textbf{77.4} & \textbf{81.4} & \textbf{85.1} 
				
			\end{tabular}
		}
	\end{center}
	\caption{Comparison vs other methods\ on the Cityscapes \textit{test} set. Best results in each class are represented in bold.}
\label{tab:table4}
\end{table*}

\subsubsection{Results on Mapillary Vistas}
Mapillary Vistas ~\cite{MVD2017} is a large dataset containing $25,000$ high resolution images annotated into $66$ object categories. For Mapillary, we used HRNet-OCR as the trunk along with our proposed multi-scale attention method. Because Mapillary images can have very high and varied resolutions, we resize the images such that the long edge is 2177 as was done in ~\cite{cheng2019panopticdeeplab}. We initialize the HRNet part of the model with weights from HRNet trained on ImageNet classification. Because of the greater memory requirements for the 66 classes in Mapillary, we decreased the crop size to 1856 x 1024. In Table \ref{tab:table5} we show results of our method on Mapillary validation set. Our single-model based method achieves $61.1$, which is $2.4$ higher than the next closest method, Panoptic Deeplab ~\cite{cheng2019panopticdeeplab}, which uses ensemble of models to achieve $58.7$. 

\begin{table}[htb!]
\centering
\begin{tabular}{|c       |c|} \hline
               Method  & mIOU \\ \hline\hline
Seamless~\cite{Porzi_2019_CVPR} & 50.4 \\ \hline               
DeeperLab~\cite{yang2019deeperlab} & 55.3 \\ \hline              
Panoptic DeepLab ~\cite{cheng2019panopticdeeplab}             &  56.8   \\ \hline
Panoptic DeepLab ( Ensemble ) ~\cite{cheng2019panopticdeeplab} &  58.7   \\ \hline
Ours                          &  \textbf{61.1}   \\ \hline
\end{tabular}
\vspace{2ex}
\caption{Comparison of results on Mapillary \textit{validation} set. Best results in each class are represented in bold.}
\label{tab:table5}
\end{table}

\section{Conclusion}
In this work, we present a hierarchical multi-scale attention approach for semantic segmentation. Our approach yields an improvement in segmentation accuracy while also being memory and computationally efficient, both of which are practical concerns. Training efficiency limits how fast research can be done while GPU memory efficiency limits how large of a crop networks can be trained with, which can also limit network accuracy. We empirically show consistent improvement in Cityscapes and Mapillary using our proposed approach.

\textbf{Acknowledgements}: We'd like to thank Sanja Fidler, Kevin Shih, Tommi Koivisto and Timo Roman for helpful discussions. 

\bibliographystyle{unsrt}  
\bibliography{thebib} 

\end{document}